\documentclass{article}

\usepackage{icml2025}

\usepackage{microtype}
\usepackage{booktabs} % for professional tables
\usepackage{amsmath, amsthm, amssymb}
\usepackage{subfig}
\usepackage{graphicx}
\usepackage{amsmath}
\usepackage{amsfonts}
\usepackage{times}
\usepackage{amsmath}
\usepackage{MnSymbol}
\usepackage{wrapfig}
\usepackage[utf8]{inputenc} 
\usepackage[T1]{fontenc}  
\usepackage{hyperref}    
\usepackage{url}      
\usepackage{booktabs}    
\usepackage{nicefrac}    
\usepackage{microtype}   
\usepackage{multirow}
\usepackage[toc,page,header]{appendix}
\usepackage{amsfonts}
\usepackage[normalem]{ulem}
\useunder{\uline}{\ul}{}

\usepackage{booktabs}
\usepackage{comment}
\usepackage{booktabs} % For better looking tables
\usepackage{siunitx}
\usepackage{colortbl}
\newtheorem{theorem}{Theorem}
\newtheorem{theorem_a}{Theorem}
\newtheorem{lemma}{Lemma}
\newtheorem{lemma_a}{Lemma}

\usepackage{amsmath,amsfonts,bm}

\def\eqref#1{equation~\ref{#1}}
\def\1{\bm{1}}

\DeclareMathAlphabet{\mathsfit}{\encodingdefault}{\sfdefault}{m}{sl}
\SetMathAlphabet{\mathsfit}{bold}{\encodingdefault}{\sfdefault}{bx}{n}

\usepackage{pifont}
\usepackage{colortbl}
\usepackage{algorithm}
\usepackage{algpseudocode}
\usepackage{graphicx}
\usepackage{booktabs}
\usepackage{hyperref}
\usepackage{url}
\definecolor{magenta}{HTML}{A02B93} 
\definecolor{SeaGreen}{HTML}{2E8B57} 
\newcommand{\hlt}[1]{ {\color{black}#1}}
\usepackage{pgfplots}
\usepackage{subcaption}
\usepackage{adjustbox}
\usepackage{hyperref} 
\newcommand{\cmark}{\textcolor{blue}{\ding{51}}} % Green Checkmark
\newcommand{\xmark}{\textcolor{red}{\ding{55}}}   % Red Crossmark
\usepackage{hyperref}

\usepackage{amsmath}
\usepackage{amssymb}
\usepackage{mathtools}
\usepackage{amsthm}
\usepackage[capitalize,noabbrev]{cleveref}

\theoremstyle{plain}

\theoremstyle{definition}

\theoremstyle{remark}

\usepackage[textsize=tiny]{todonotes}

\begin{document}

\twocolumn[
\icmltitle{FLAMES: A Hybrid Spiking-State Space Model for Adaptive Memory Retention in Event-Based Learning}

\icmlsetsymbol{equal}{*}

\begin{icmlauthorlist}
\icmlauthor{Biswadeep Chakraborty}{gatech}
\icmlauthor{Saibal Mukhopadhyay}{gatech}
\end{icmlauthorlist}

\icmlaffiliation{gatech}{Department of Electrical and Computer Engineering, Georgia Institute of Technology, Atlanta, GA, USA}

\icmlcorrespondingauthor{Biswadeep Chakraborty}{biswadeep@gatech.edu}

\icmlkeywords{Spiking Neural Networks, State Space Models, Machine Learning, Neuromorphic Computing}

\vskip 0.3in
]

\printAffiliationsAndNotice{\icmlEqualContribution}

\begin{abstract}
   We propose \textbf{FLAMES (Fast Long-range Adaptive Memory for Event-based Systems)}, a novel hybrid framework integrating structured state-space dynamics with event-driven computation. At its core, the \textit{Spike-Aware HiPPO (SA-HiPPO) mechanism} dynamically adjusts memory retention based on inter-spike intervals, preserving both short- and long-range dependencies. To maintain computational efficiency, we introduce a normal-plus-low-rank (NPLR) decomposition, reducing complexity from $\mathcal{O}(N^2)$ to $\mathcal{O}(Nr)$.  FLAMES achieves state-of-the-art results on the Long Range Arena benchmark and event datasets like HAR-DVS and Celex-HAR. By bridging neuromorphic computing and structured sequence modeling, FLAMES enables scalable long-range reasoning in event-driven systems.
\end{abstract}

\section{Introduction}

\begin{figure*}
  \begin{center}
    \includegraphics[width=0.89\textwidth]{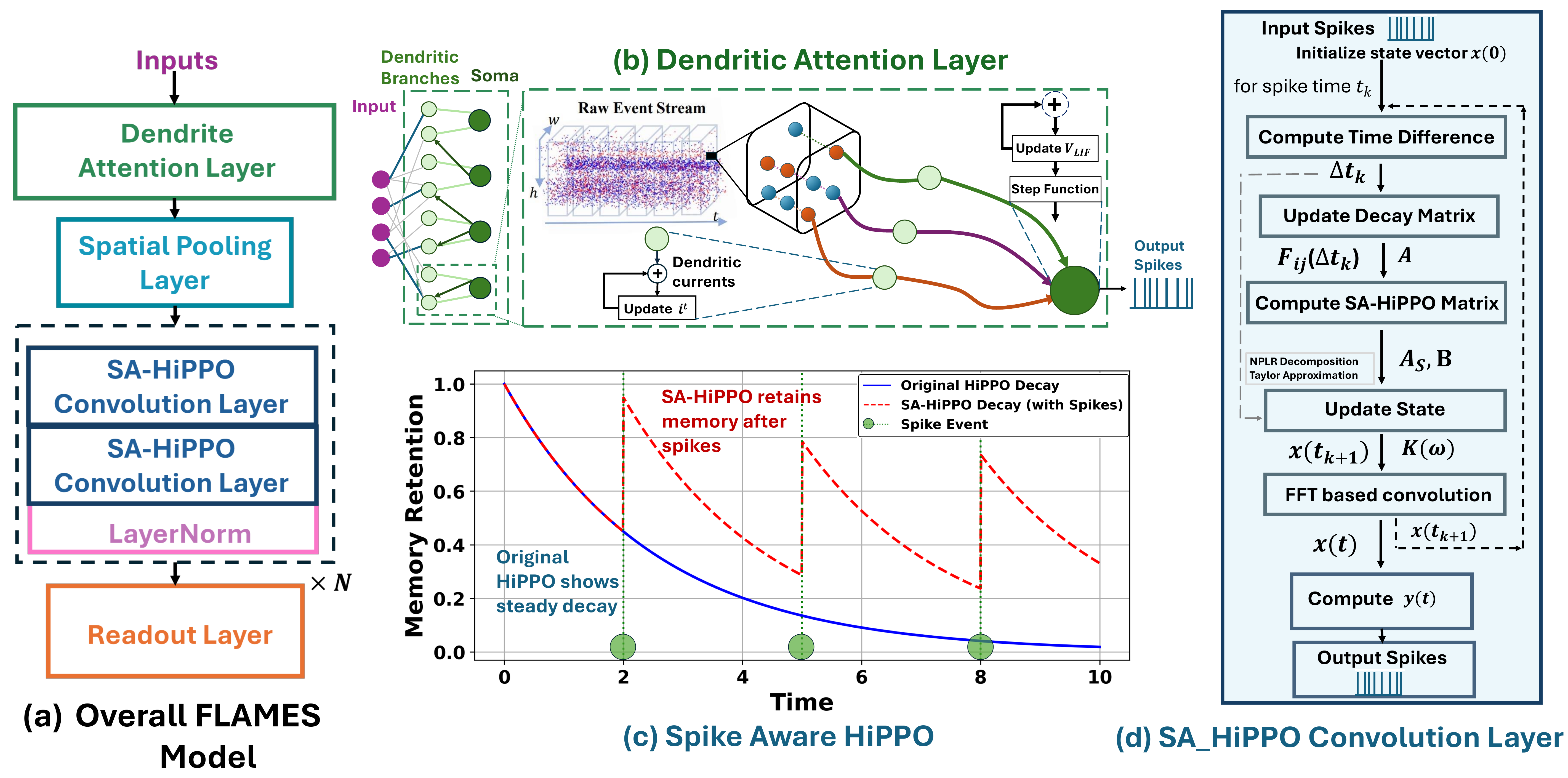}
  \end{center}
    \caption{Block diagram of the proposed model architecture. Input spikes are processed event-by-event} by the Dendrite Attention Layer, which extracts spatio-temporal features from local dendritic branches. The Spatial Pooling Layer aggregates spikes, followed by SA-HiPPO Convolution and LayerNorm layers, which are repeated \( N \) times to enable hierarchical feature extraction and long-range temporal dependency modeling.
  \label{fig:overall}
\end{figure*}

Spiking Neural Networks (SNNs) offer a biologically inspired and computationally efficient framework for processing temporal data through sparse, event-driven computations. Unlike conventional artificial neural networks, which rely on dense, synchronous activations, SNNs operate on discrete spike events, making them particularly effective for neuromorphic vision and real-time sensor processing \cite{roy2019towards, davies2018loihi, rathi2023exploring}. This event-driven paradigm enables significant energy savings by performing computation only when information is available. However, a fundamental limitation of SNNs is their difficulty in retaining long-range temporal dependencies, restricting their applicability to complex sequence modeling tasks.

The core of this limitation stems from the temporal processing characteristics of standard Leaky Integrate-and-Fire (LIF) neurons, where information is stored in exponentially decaying membrane potentials \cite{bengio1994learning, hochreiter1997long}. While this mechanism allows for efficient local integration of spike events, it also leads to rapid memory loss, severely limiting an SNN’s ability to capture dependencies over long timescales \cite{neftci2019surrogate, bellec2018long}. Approaches that extend SNN memory through recurrent architectures \cite{wu2018spatio, shrestha2018slayer, zheng2024temporal} introduce dense state representations that compromise the inherent sparsity of spike-based computation. This presents a trade-off: increasing memory retention in SNNs typically comes at the cost of increased computational complexity.

State Space Models (SSMs) have recently emerged as an effective alternative for long-range sequence modeling \cite{gu2020hippo, gu2022efficiently}. A foundational component of these models is the \textbf{HiPPO} (High-Order Polynomial Projection Operator) framework \cite{gu2020hippo}, which optimally compresses and updates memory states over time. HiPPO constructs a set of orthogonal basis functions to continuously project and retain historical information, enabling efficient long-range memory retention in a structured manner. However, HiPPO and its extensions (e.g., S4 \cite{gu2022efficiently}) are designed for dense, continuous inputs, making them fundamentally incompatible with the sparse and asynchronous nature of SNNs.

To bridge this gap, we introduce \textbf{FLAMES (Fast Long-range Adaptive Memory for Event-based Systems)}, a framework that integrates event-driven computation in SNNs with SSM-inspired structured memory to enable efficient long-range temporal modeling. The key innovation in FLAMES is the \textbf{Spike-Aware HiPPO (SA-HiPPO) mechanism}, which dynamically scales memory retention based on inter-spike intervals. Unlike the baseline HiPPO, which assumes continuous input and updates memory uniformly over time, SA-HiPPO adapts the state-space updates to the event-driven nature of SNNs. This adaptation ensures that memory is retained selectively—short-term dependencies are reinforced during frequent spiking, while long-term dependencies are preserved over sparse spiking intervals. By directly addressing the event-driven constraints of SNNs, FLAMES achieves efficient memory retention without sacrificing the sparsity and computational efficiency inherent to spike-based processing.

FLAMES builds on three core theoretical innovations. First, we develop a spike-aware state transition mechanism that dynamically modulates memory retention based on inter-spike intervals \cite{gu2022train, lotfi2020long}, ensuring stability while maintaining computational efficiency. Second, we introduce a normal-plus-low-rank (NPLR) decomposition specifically optimized for sparse updates \cite{gu2022efficiently}, reducing computational complexity from $\mathcal{O}(N^2)$ to $\mathcal{O}(Nr)$ while preserving expressivity. Third, we establish theoretical bounds on the trade-off between memory capacity and computational cost \cite{wei2024event, chakraborty2023heterogeneous_frontiers}, providing a rigorous framework for optimizing event-driven memory representations.

Our key contributions are:
\begin{itemize}
    \item \textbf{Spike-Aware HiPPO (SA-HiPPO):} A dynamic memory mechanism that adjusts state retention based on inter-spike intervals, preserving both short- and long-range dependencies.
    \item \textbf{Computationally Efficient State Updates:} A normal-plus-low-rank (NPLR) decomposition reduces update complexity from $\mathcal{O}(N^2)$ to $\mathcal{O}(Nr)$, enabling scalable state-space modeling in event-driven systems.
    \item \textbf{Hybrid Spiking-State Space Framework:} Combines the sparsity of SNNs with structured memory updates, enabling efficient processing of long-range dependencies.
    \item \textbf{Dendritic Attention Layer:} Extends LIF neurons with multi-timescale dendritic processing, enhancing feature extraction from sparse event-driven data.
    \item \textbf{Comprehensive Empirical Validation:} FLAMES achieves state-of-the-art results across diverse benchmarks, including long-range dependency modeling (LRA), gesture recognition (DVS Gesture), and human activity recognition (HAR-DVS, Celex-HAR) demonstrating efficiency and scalability in event-driven tasks.
\end{itemize}

% Our work establishes a theoretical bridge between the efficiency of event-driven computation \cite{martin2024alert, ren2024rethinking} and the expressivity of state space models \cite{gu2022efficiently}. By proving that spike-based architectures can efficiently model long-range dependencies without compromising their fundamental advantages, FLAMES opens new possibilities for neuromorphic computing in domains requiring both computational efficiency and sophisticated temporal reasoning.

\begin{table*}[htbp]
\centering
\caption{Comparison of FLAMES with prior methods, highlighting features like memory retention, event-driven processing, scalability, efficiency, asynchronous updates, and adaptability.}
\label{tab:FLAMES_comparison_novelty}
\small
\resizebox{0.9\textwidth}{!}{%
\begin{tabular}{@{}>{\raggedright\arraybackslash}p{3.5cm} c c c c c c c@{}}
\toprule
\textbf{Model} & \textbf{Type} & \textbf{\begin{tabular}[c]{@{}c@{}}Dynamic \\ Memory \\ Retention\end{tabular}} & \textbf{\begin{tabular}[c]{@{}c@{}}Event-Driven \\ Processing\end{tabular}} & \textbf{\begin{tabular}[c]{@{}c@{}}Scalable \\ Long-Range \\ Modeling\end{tabular}} & \textbf{\begin{tabular}[c]{@{}c@{}}Low \\ Computational \\ Overhead\end{tabular}} & \textbf{\begin{tabular}[c]{@{}c@{}}Fully \\ Asynchronous \\ Updates\end{tabular}} & \textbf{\begin{tabular}[c]{@{}c@{}}Adaptability \\ to Sparse \\ Data\end{tabular}} \\
\midrule
SpikingLMU \cite{liu2024lmuformer} & SSM & \cmark & \xmark & \xmark & \xmark & \xmark & \cmark \\
BinaryS4D \cite{stan2024learning} & SSM & \xmark & \xmark & \cmark & \xmark & \xmark & \xmark \\
HiPPO \cite{gu2020hippo} & SSM & \xmark & \xmark & \cmark & \xmark & \xmark & \xmark \\
DH-LIF \cite{zheng2024temporal} & SNN & \xmark & \cmark & \xmark & \xmark & \cmark & \xmark \\
EventMamba \cite{ren2024rethinking} & Hybrid CNN-SSM & \xmark & \xmark & \cmark & \xmark & \xmark & \xmark \\
EventNet \cite{martin2024alert} & Transformer & \xmark & \xmark & \cmark & \xmark & \cmark & \xmark \\
SpikeRWKV \cite{yao2024spike} & Transformer & \xmark & \cmark & \cmark & \xmark & \cmark & \cmark \\
S4 \cite{gu2022efficiently} & SSM & \xmark & \xmark & \cmark & \cmark & \xmark & \xmark \\
\rowcolor[HTML]{EAF7EA}
FLAMES (Ours) & SSM-SNN & \cmark & \cmark & \cmark & \cmark & \cmark & \cmark \\
\bottomrule
\end{tabular}
}
\vspace{-4mm}
\end{table*}

\section{Related Works}
\textbf{State Space Models for Temporal Processing: }
State Space Models (SSMs) have emerged as a powerful framework for modeling long-range dependencies in temporal data \cite{gu2020hippo, gu2022efficiently}. One foundational approach is the \textbf{HiPPO} (High-Order Polynomial Projection Operator) framework \cite{gu2020hippo}, which provides a theoretically optimal way to compress and propagate memory in continuous-time recurrent models. HiPPO constructs a set of basis functions that maintain information over time using structured state-space dynamics, offering efficient sequence modeling. However, a key limitation of HiPPO and its extensions (e.g., S4 \cite{gu2022efficiently}, S5 \cite{smith2022simplified}) is that they assume dense, synchronous inputs, making them inherently incompatible with the sparse, event-driven nature of Spiking Neural Networks (SNNs) \cite{gu2021combining, hasani2021liquid, chakraborty2024topological}. 

Recent efforts have attempted to bridge this gap by integrating SSMs into spiking architectures. SpikingLMU \cite{liu2024lmuformer} combines Legendre Memory Units (LMUs) with SNNs to retain long-range dependencies, but it requires additional recurrent structures, increasing computational overhead. BinaryS4D \cite{stan2024learning} leverages a binarized version of S4 in spiking networks but still relies on dense matrix multiplications, reducing the efficiency of event-driven processing. These approaches highlight the fundamental challenge of adapting structured memory mechanisms to neuromorphic computing without sacrificing computational efficiency \cite{chakraborty2023braindate, kang2024learning}.

\textbf{Event-Driven Processing and Memory in SNNs: }
Event-driven processing remains a key challenge in neuromorphic computing \cite{schuman2022opportunities, gallego2020event}. While models like SLAYER \cite{shrestha2018slayer} and surrogate gradient methods \cite{neftci2019surrogate, bellec2018long} improve training stability, they struggle with long-range dependencies due to rapid membrane potential decay \cite{wu2018spatio}. To enhance memory in SNNs, recent works have incorporated dendritic processing, such as DH-LIF \cite{zheng2024temporal} and GLIF \cite{yao2022glif}, which introduce multi-compartment neuron models. However, these methods still lack structured memory mechanisms like those found in SSMs.

Alternative approaches involve frame-based and hybrid event-adapted architectures, including transformers \cite{vaswani2017attention}, which model long-range dependencies but are computationally expensive for sparse event-driven data. Event-adapted models like Alert-Transformer \cite{martin2024alert} and SpikeRWKV \cite{yao2024spike} attempt to balance sparsity and sequence modeling but remain fundamentally limited in scaling to long-horizon dependencies while preserving event-driven efficiency \cite{chakraborty2024exploiting}, \cite{chakraborty2023braindate}, \cite{kumawat2024robokoop}.

\textbf{Adaptive Memory in SNNs: }
Memory retention in SNNs has been explored through recurrent spiking networks and biologically motivated adaptation mechanisms \cite{bellec2018long, zenke2021brain, chakraborty2023heterogeneous_frontiers}. Liquid Time-Constant Networks (LTCs) \cite{hasani2021liquid} provide temporal adaptivity by adjusting neuron dynamics, but they incur significant computational overhead. Other methods explore learning-based adaptations, such as STDP variants \cite{chakraborty2021characterization} and spike-aware architecture search \cite{chakraborty2021mu}. Our proposed FLAMES framework builds upon these ideas by introducing \textbf{SA-HiPPO}, which dynamically aligns state-space memory updates with inter-spike intervals. Unlike standard HiPPO-based methods, which assume continuous and dense input streams, SA-HiPPO explicitly models memory retention in an event-driven setting \cite{wei2024event, gu2022train}. By leveraging structured state-space updates while maintaining sparse computation, FLAMES enables efficient long-range modeling for neuromorphic vision \cite{gehrig2024low}, temporal reasoning tasks \cite{xiao2024temporal}, and emerging low-power edge applications \cite{chakraborty2024sparse, chakraborty2023brainijcnn, kumawat2024stage, chakraborty2024dynamical}.

\section{Methods}

FLAMES is designed to combine the best of spiking and continuous representations. It processes inputs in an event-driven manner, maintaining sparse computation in early layers. However, after spatial pooling, it transitions into a structured state-space model, allowing for more efficient long-range dependency modeling. This hybrid approach retains the advantages of SNNs while overcoming their limitations in memory retention. As shown in Figure \ref{fig:overall}, our model consists of five key components: (1) a \textbf{Dendrite Attention Layer} for multi-scale temporal feature extraction, (2) a \textbf{Spatial Pooling Layer} for dimensionality reduction, (3) the core \textbf{SA-HiPPO Convolution Layer} with our novel \textit{SA-HiPPO} mechanism, (4) \textbf{Layer Normalization}, and (5) a \textbf{Readout Layer} for downstream tasks. These components work together to enable efficient processing of long-range temporal dependencies while preserving the computational benefits of spike-based processing.

\textbf{Input \& Output Processing: }The model processes input as a sequence of spike events, each defined by $(x, y, t, p)$, where $(x, y)$ represents spatial coordinates, $t$ is the timestamp, and $p$ indicates polarity. This event-driven representation maintains the sparse, asynchronous nature of neuromorphic data, enabling efficient processing.
For the output processing, we use a two-step processing strategy: first, a layer normalization technique stabilizes training by reducing activation variability through centering and scaling, and then an event-pooling mechanism subsamples the temporal sequence. This event-pooling approach intelligently selects the most relevant temporal features by averaging activations over discrete intervals, followed by a linear transformation that maps the pooled representation to the final output. By dynamically reducing computational complexity while preserving critical temporal information, this approach enables efficient and scalable processing of long event sequences across different downstream tasks.

\subsection{Dendrite Attention Layer: } The model begins by passing the input through the \textit{Dendrite Attention Layer}, constructed using DH-LIF neurons \cite{Zheng2024}, as shown in Figure \ref{fig:overall}(b). Each DH-LIF neuron has multiple dendritic branches, each characterized by a different timing factor \( \tau_d \), enabling it to capture temporal dynamics across various scales. This is essential for accommodating the diverse timescales present in asynchronous spike inputs. 

The dynamics of dendritic current \( \mathbf{i}_d(t) \) are governed by:
\[
\mathbf{i}_d(t+1) = \alpha_d \mathbf{i}_d(t) + \sum_{j \in \mathcal{N}_d} \mathbf{w}_j p_j,
\]
where \( \alpha_d = e^{-\frac{1}{\tau_d}} \) is the decay rate for branch \( d \), and \( \mathbf{w}_j \) represents the synaptic weight associated with presynaptic input \( p_j \). The set \( \mathcal{N}_d \) represents the presynaptic inputs connected to dendrite \( d \), ensuring that each dendrite captures temporal features independently, acting as a temporal filter. Unlike a standard CUBA LIF neuron model, which integrates all inputs uniformly at the soma with a single timescale, the dendritic attention layer introduces multiple dendritic branches, each independently filtering inputs at different temporal scales. This design enables the neuron to selectively process asynchronous inputs and retain information across diverse temporal windows, providing greater flexibility and adaptability.

The dendritic currents from each branch are aggregated at the soma of the LIF neuron, resulting in the membrane potential:
\[
V(t+1) = \beta V(t) + \sum_{d} \mathbf{g}_d \mathbf{i}_d(t),
\]
where \( \beta = e^{-\frac{1}{\tau_s}} \) represents the soma’s decay rate, and \( \mathbf{g}_d \) represents the coupling strength of dendrite \( d \) to the soma. A spike is generated whenever the membrane potential exceeds a threshold \( V_{\text{th}} \), allowing the neuron to selectively fire only when sufficiently excited.

\subsection{Spatial Pooling Layer}
Event-based vision sensors often generate high-dimensional spatial data (e.g., 1280×800 in CeleX), creating computational bottlenecks in processing. To address this, we introduce a spatially-adaptive pooling layer that reduces dimensionality while preserving critical temporal information in the spike streams.

Given an input spike activity \(I(x, y, t)\) at spatial location \((x, y)\) and time \(t\), the pooling operation is defined as:
\[
I_{\text{pooled}}(x', y', t) = \max_{(x, y) \in P(x', y')} \Big\{ I(x, y, t) \Big\},
\]
where \(P(x', y')\) represents a dynamic pooling window centered at \((x', y')\). Unlike traditional frame-based pooling, this operation:
1. Maintains precise spike timing by operating independently at each timestep
2. Preserves spatial locality of events through maximum pooling
3. Reduces the spatial feature dimension by a factor of \(k^2\) for a \(k \times k\) pooling window

This spatial compression is particularly critical for high-resolution event cameras where the raw spatial dimension can exceed 1M pixels, enabling efficient downstream processing while retaining the essential spatio-temporal structure of the event stream.

\subsection{SA-HiPPO Convolution Layer}

The SA-HiPPO (Spike-Aware HiPPO) mechanism extends the standard HiPPO framework to efficiently model long-range dependencies in sparse, event-driven data. Unlike conventional spiking architectures that struggle with temporal retention, SA-HiPPO dynamically adjusts its state transition dynamics based on inter-spike intervals, preserving relevant information while maintaining computational efficiency.

Let $\mathbf{x}(t) \in \mathbb{R}^N$ denote the internal state vector that encodes memory at time $t$. The continuous-time dynamics of the system are governed by

\begin{equation}
\dot{\mathbf{x}}(t) = \mathbf{A}_S\mathbf{x}(t) + \mathbf{B}\mathbf{S}(t),
\label{eq:continuous_state_update}
\end{equation}

where $\mathbf{S}(t) \in \mathbb{R}^M$ represents the input spike train modeled as $ S_i(t) = \sum_k \delta(t - t_i^k),$ where $\delta(\cdot)$ is the Dirac delta function, and $t_i^k$ denotes the spike times of neuron $i$. The matrix $\mathbf{B} \in \mathbb{R}^{N \times M}$ governs how spike inputs influence state updates. 

\textbf{Adaptive State-Space Representation.} The key component of SA-HiPPO is the adaptive state transition matrix:

\begin{equation}
\mathbf{A}_S = \mathbf{A} \circ \mathbf{F}(\Delta t),
\label{eq:adaptive_state_matrix}
\end{equation}

where $\mathbf{A} \in \mathbb{R}^{N \times N}$ is the base HiPPO matrix responsible for structured memory compression, and $\mathbf{F}(\Delta t) \in \mathbb{R}^{N \times N}$ is a decay matrix that modulates memory retention as a function of inter-spike intervals $\Delta t$. The operator $\circ$ denotes the Hadamard (element-wise) product, ensuring that each state dimension decays independently. The decay matrix $\mathbf{F}(\Delta t)$ is defined as

\begin{equation}
F_{ij}(\Delta t) = e^{-\alpha_{ij} \Delta t},
\end{equation}

where $\alpha_{ij} > 0$ are learnable parameters that control the rate at which memory decays between state components $i$ and $j$. This formulation allows the system to dynamically adjust its memory retention. When spikes occur frequently ($\Delta t$ is small), $F_{ij}(\Delta t) \approx 1$, preserving short-term dependencies. Conversely, when spikes are sparse ($\Delta t$ is large), $F_{ij}(\Delta t) \to 0$, facilitating memory reset.

\textbf{Discrete-Time State Update.} In an event-driven setting, state updates occur asynchronously at spike arrival times. Given an inter-spike interval $\Delta t_k = t_{k+1} - t_k$, the discrete-time state update follows

\begin{equation}
\mathbf{x}(t_{k+1}) = e^{\mathbf{A}_S\Delta t_k} \mathbf{x}(t_k) + \mathbf{A}_S^{-1} (e^{\mathbf{A}_S\Delta t_k} - \mathbf{I}) \mathbf{B} \mathbf{S}(t_k).
\label{eq:discrete_state_update}
\end{equation}

The term $e^{\mathbf{A}_S \Delta t_k} \mathbf{x}(t_k)$ determines the natural decay of past memory over the interval $\Delta t_k$, while the second term accounts for the influence of incoming spikes on the memory state. For computational efficiency, the matrix exponential is approximated using a second-order Taylor expansion:

\begin{equation}
e^{\mathbf{A}_S\Delta t} \approx \mathbf{I} + \mathbf{A}_S\Delta t + \frac{(\mathbf{A}_S\Delta t)^2}{2}.
\end{equation}

This approximation reduces computational overhead while preserving essential state dynamics.

\textbf{Memory Retention and Stability.} The memory retention properties of SA-HiPPO are determined by the eigenvalues of $\mathbf{A}_S$. Since $\mathbf{A}_S = \mathbf{A} \circ \mathbf{F}(\Delta t)$ and $\mathbf{A}$ is a Hurwitz matrix, the system remains stable while dynamically adjusting its memory horizon. 

For any input spike train $\mathbf{S}(t)$ with bounded inter-spike intervals $\Delta t \leq T$, the system's state norm satisfies

\begin{equation}
\|\mathbf{x}(t)\| \leq e^{-\alpha t}\|\mathbf{x}_0\| + \frac{\|\mathbf{B}\|S_\infty}{\alpha} (1 - e^{-\alpha t}),
\end{equation}

where $\alpha = \min_i |\text{Re}(\lambda_i)| > 0$ represents the slowest decay mode of $\mathbf{A}_S$, and $S_\infty$ is an upper bound on the spike magnitude. This bound ensures that long-range dependencies are preserved while controlling memory decay.

\textbf{Illustration: Impact of Adaptive Decay on Memory.} Consider two neurons generating spikes at different rates. Suppose neuron A fires regularly every 5ms while neuron B fires sporadically with intervals between 10–20ms. For neuron A (frequent spiking), the decay matrix maintains relatively stable memory retention $\displaystyle F_{ij}(5\text{ms}) \approx e^{-5\alpha_{ij}}.$

For neuron B (irregular spiking), the decay adapts dynamically. During long intervals (20ms), the decay is stronger, $F_{ij}(20\text{ms}) \approx e^{-20\alpha_{ij}}$, leading to faster forgetting of older information. Conversely, during shorter intervals (10ms), the decay is weaker, $F_{ij}(10\text{ms}) \approx e^{-10\alpha_{ij}}$, preserving more recent temporal patterns. This adaptive behavior enables SA-HiPPO to efficiently process both regular and irregular spike patterns while maintaining relevant long-range dependencies.

\section{Theoretical Analysis}

This section establishes theoretical guarantees for our adaptive state space framework, focusing on three key aspects: computational efficiency, temporal dependency preservation, and stability. We present formal bounds and analyze how different components interact to ensure robust and efficient processing. \textbf{\textit{Detailed proofs are provided in the supplementary material.}}

\subsection{Computational Complexity}

\begin{lemma}[Computational Efficiency of Adaptive State Updates]
For the state space model:
\[
\dot{\mathbf{x}}(t) = \mathbf{A} \cdot \mathbf{x}(t) + \mathbf{B} \cdot \mathbf{S}(t),
\]
where $\mathbf{x}(t) \in \mathbb{R}^N$, $\mathbf{A} \in \mathbb{R}^{N \times N}$, and $\mathbf{S}(t) \in \mathbb{R}^M$ represents spike inputs, the computational complexity for state updates at each spike event is $O(N^2)$, and reduces to $O(Nr)$ with our NPLR decomposition, where $r \ll N$.
\end{lemma}

This result quantifies the computational advantage gained through our architectural optimizations. The NPLR decomposition significantly reduces computational overhead while maintaining model expressivity, making it suitable for real-time processing of event streams.

\subsection{Long-Range Memory Analysis}

\begin{theorem}[Temporal Dependency Preservation]
Let $\mathbf{x}(t) \in \mathbb{R}^N$ evolve according to our adaptive state space dynamics:
\[
\dot{\mathbf{x}}(t) = \mathbf{A} \cdot \mathbf{x}(t) + \mathbf{B} \cdot \mathbf{S}(t),
\]
where $\mathbf{A}$ is a HiPPO matrix with eigenvalues satisfying $\text{Re}(\lambda_i) < 0$, $\|\mathbf{S}(t)\| \leq S_{\infty}$, and $\mathbf{x}(0) = \mathbf{x}_0$. Then:
\[
\|\mathbf{x}(t)\| \leq e^{-\alpha t}\|\mathbf{x}_0\| + \frac{\|\mathbf{B}\|S_{\infty}}{\alpha}(1 - e^{-\alpha t}),
\]
where $\alpha = \min_i|\text{Re}(\lambda_i)| > 0$ controls memory retention.
\end{theorem}

This theorem establishes that our framework effectively balances the preservation of long-range dependencies with the decay of outdated information. The memory retention factor $\alpha$ provides a tunable parameter for controlling this trade-off.

\begin{table*}[h]
\centering
\caption{Results comparing the accuracy of our model against state-of-the-art architectures on LRA benchmark tasks.}
\label{tab:lra}
\resizebox{0.8\textwidth}{!}{%

\begin{tabular}{|l|c|c|c|c|c|c|}
\hline
\textbf{Model} & \textbf{SNN} & \textbf{ListOps} & \textbf{Text} & \textbf{Retrieval} & \textbf{Image} & \textbf{Pathfinder} \\ \hline
S4 (Original) \cite{gu2022efficiently} & No & 58.35 & 76.02 & 87.09 & 87.26 & 86.05 \\ 
S4 (Improved) \cite{gu2022efficiently} & No & 59.60 & 86.82 & 90.90 & 88.65 & 94.20 \\ 
Transformer \cite{vaswani2017attention} & No & 36.37 & 64.27 & 57.46 & 42.44 & 71.40 \\ 
Sparse Transformer \cite{tay2020long} & No & 17.07 & 63.58 & 59.59 & 44.24 & 71.71 \\ 
Linformer \cite{wang2020linformer} & No & 35.70 & 53.94 & 52.27 & 38.56 & 76.34 \\ 
Linear Transformer \cite{tay2020long} & No & 16.13 & 65.90 & 53.09 & 42.34 & 75.30 \\ 
FLASH-quad \cite{hua2022transformer} & No & 42.20 & 64.10 & 83.00 & 48.30 & 83.62 \\ 
Spiking LMUFormer \cite{liu2024lmuformer} & Yes & 37.30 & 65.80 & 79.76 & 55.65 & 72.68 \\ 
TransNormer T2 \cite{qin2022devil} & No & 41.60 & 72.20 & 83.82 & 49.60 & 76.60 \\ 
BinaryS4D \cite{stan2024learning} & Partial & 54.80 & 82.50 & 85.30 & 82.00 & 82.60 \\ 
\textbf{FLAMES (Our Model)} & Yes & \textbf{59.08} & \textbf{79.41} & \textbf{89.62} & \textbf{79.88} & \textbf{86.47} \\ \hline
\end{tabular}%
}
\end{table*}

\subsection{Numerical Stability}

\begin{lemma}[State Update Stability]
For the truncated Taylor expansion approximation of the matrix exponential:
\[
e^{\mathbf{A}\Delta t} \approx \mathbf{I} + \mathbf{A}\Delta t + \frac{\mathbf{A}^2\Delta t^2}{2!} + \cdots + \frac{\mathbf{A}^n\Delta t^n}{n!},
\]
the approximation error $\mathbf{E}_n$ is bounded by:
\[
\|\mathbf{E}_n\| \leq \frac{\|\mathbf{A}\Delta t\|^{n+1}}{(n+1)!}.
\]
\end{lemma}

This bound guides the practical implementation of state updates, allowing us to choose the minimum number of terms needed for a desired accuracy level while maintaining computational efficiency.

\begin{theorem}[Global Stability]
The state trajectory $\mathbf{x}(t)$ of our framework remains bounded ($\|\mathbf{x}(t)\| \leq C$ for some $C > 0$) under the following conditions:
\begin{enumerate}
    \item Bounded inputs: $\|\mathbf{S}(t)\| \leq S_{\infty}$ for all $t \geq 0$
    \item Hurwitz stability: All eigenvalues of $\mathbf{A_S}$ have negative real parts
    \item Lyapunov condition: There exists positive definite $\mathbf{P}$ satisfying $\mathbf{A_S}^T\mathbf{P} + \mathbf{P}\mathbf{A_S} = -\mathbf{Q}$
\end{enumerate}
\end{theorem}

This result guarantees the framework's stability for continuous operation, ensuring that the state representation remains well-behaved even during extended processing of event streams.

\subsection{Efficiency-Memory Trade-offs}
Our theoretical analysis reveals key trade-offs between computational efficiency and memory capacity. The NPLR decomposition reduces complexity from $O(N^2)$ to $O(Nr)$ while maintaining expressive power through the low-rank structure. Similarly, the memory retention factor $\alpha$ provides a tunable parameter for balancing the preservation of temporal dependencies with computational efficiency. These trade-offs guide the practical deployment of our framework in resource-constrained environments.

\section{Experiments and Results}

\begin{figure*} 
\centering 
\includegraphics[width=\linewidth]{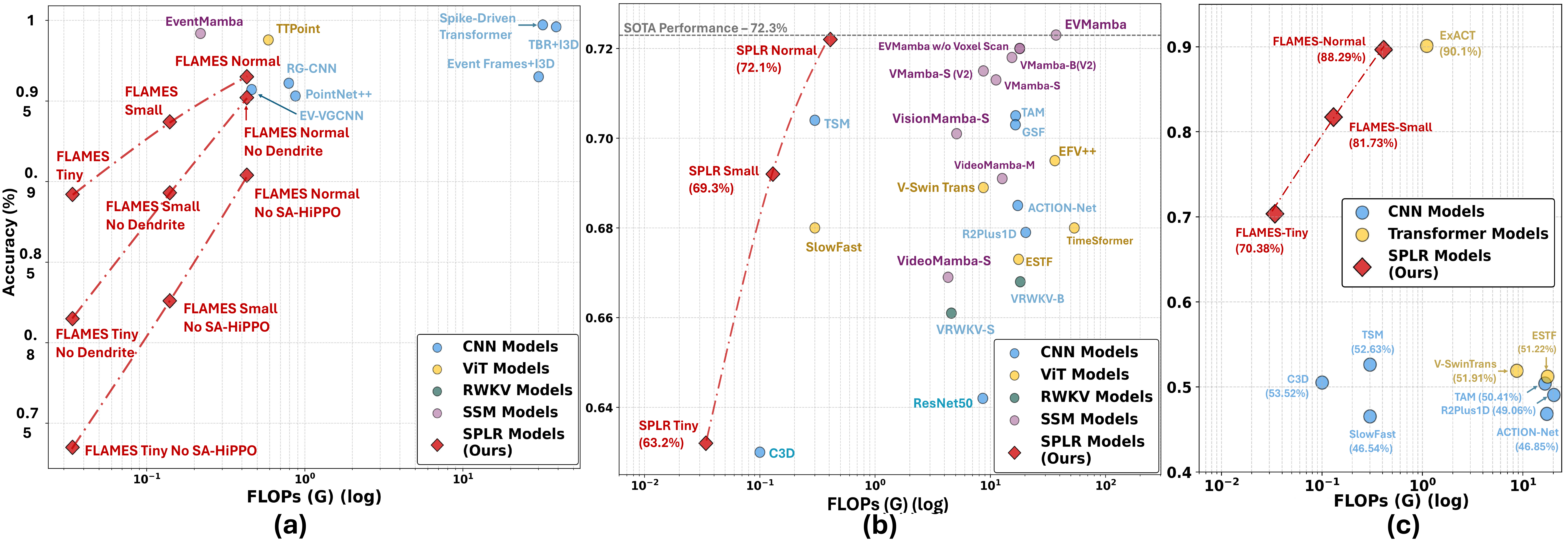} 
\caption{ Accuracy vs. FLOPS (G) on (a) \textit{DVSGesture128} , (b) \textit{Celex-HAR} and (c) HAR-DVS \cite{wang2024hardvs} datasets comparing FLAMES variants with other SOTA models. Figure (a) shows the ablation studies showing the impact of removing the Dendrite Attention Layer or replacing SA-HiPPO with standard LIF neurons. Note: There are no spike-based designs for Celex-HAR}
\label{fig:flops} 
\end{figure*}

\textbf{Experimental Setup: } We evaluate the FLAMES model on a variety of datasets to demonstrate its effectiveness in processing asynchronous, event-driven data. For all experiments, the FLAMES model processes inputs on an event-by-event basis, dynamically updating its hidden state with each incoming spike. This approach preserves high temporal resolution and captures fine-grained spatio-temporal dependencies without accumulating events into frames. Below, we summarize the experimental setup for the primary datasets. Details for additional datasets, including Sequential CIFAR-10 and CIFAR-100 \cite{krizhevsky2009learning}, SHD, and SSC \cite{cramer2020heidelberg}, is given in Suppl. Sec. \ref{sec:experiments}.
% \begin{figure}
%   \begin{center}
%     \includegraphics[width=0.8\columnwidth]{Figures/har_dvs_new.pdf}
%   \end{center}
%   \caption{Accuracy vs. FLOPS (G) on HAR-DVS Dataset\cite{wang2024hardvs}}
%   \label{fig:hardvs}
%   % \vspace{-5mm}
% \end{figure} 
\textit{DVS Gesture Dataset \cite{amir2017low}:} Contains event streams of 11 hand gestures from 29 subjects recorded with a Dynamic Vision Sensor. The FLAMES model processes real-time spikes, capturing temporal dynamics for accurate gesture classification.

\textit{HAR-DVS Dataset \cite{wang2024hardvs}:} Comprises event streams of six human activities, including walking and running, with spatial coordinates, timestamps, and polarity. FLAMES dynamically handles these sparse streams to enable real-time classification of complex activities.

\textit{Celex-HAR Dataset \cite{wang2024event}:} Utilizes high-resolution CeleX event streams of actions such as sitting and walking. The FLAMES model updates its state with each event, effectively modeling fine-grained temporal structures.

\textit{Long Range Arena (LRA) \cite{tay2020long}:} Serves as a benchmark for long-range dependency modeling. Tasks like ListOps and Path-X are transformed into event-driven formats, with FLAMES sequentially processing tokens to capture extended temporal dependencies.

% For all datasets, the event-by-event processing approach of FLAMES minimizes latency, avoids unnecessary data accumulation, and ensures efficient spatio-temporal modeling. This design makes the model particularly suited for real-time neuromorphic applications.

\textbf{Long-Range Dependencies: }We evaluate the ability of the proposed \textit{FLAMES} model to capture long-range dependencies using the \textbf{Long Range Arena (LRA)} dataset \cite{tay2020long}. The LRA benchmark evaluates models on tasks requiring long-context understanding, where Transformer-based non-spiking models often exhibit suboptimal performance due to the computational overhead of attention mechanisms, which scales poorly with increasing sequence lengths. As shown in Table \ref{tab:lra}, we benchmark our method against state-of-the-art alternatives, including SpikingLMUFormer \cite{liu2024lmuformer}, and the BinaryS4D model \cite{stan2024learning}. While BinaryS4D is not fully spiking—it relies on floating-point MAC operations for matrix multiplications—it incorporates LIF neurons to spike from an underlying SSM, providing a hybrid approach to handling long-range dependencies.

% \section{New Results }

\setlength{\textfloatsep}{2pt} % Reduce space above and below figures
\setlength{\intextsep}{2pt}    % Reduce space above and below inline text
\setlength{\abovecaptionskip}{1pt} % Reduce space above the caption
\setlength{\belowcaptionskip}{1pt} % Reduce space below the caption

\begin{figure}
  \begin{center}
    \includegraphics[width=0.8\columnwidth]{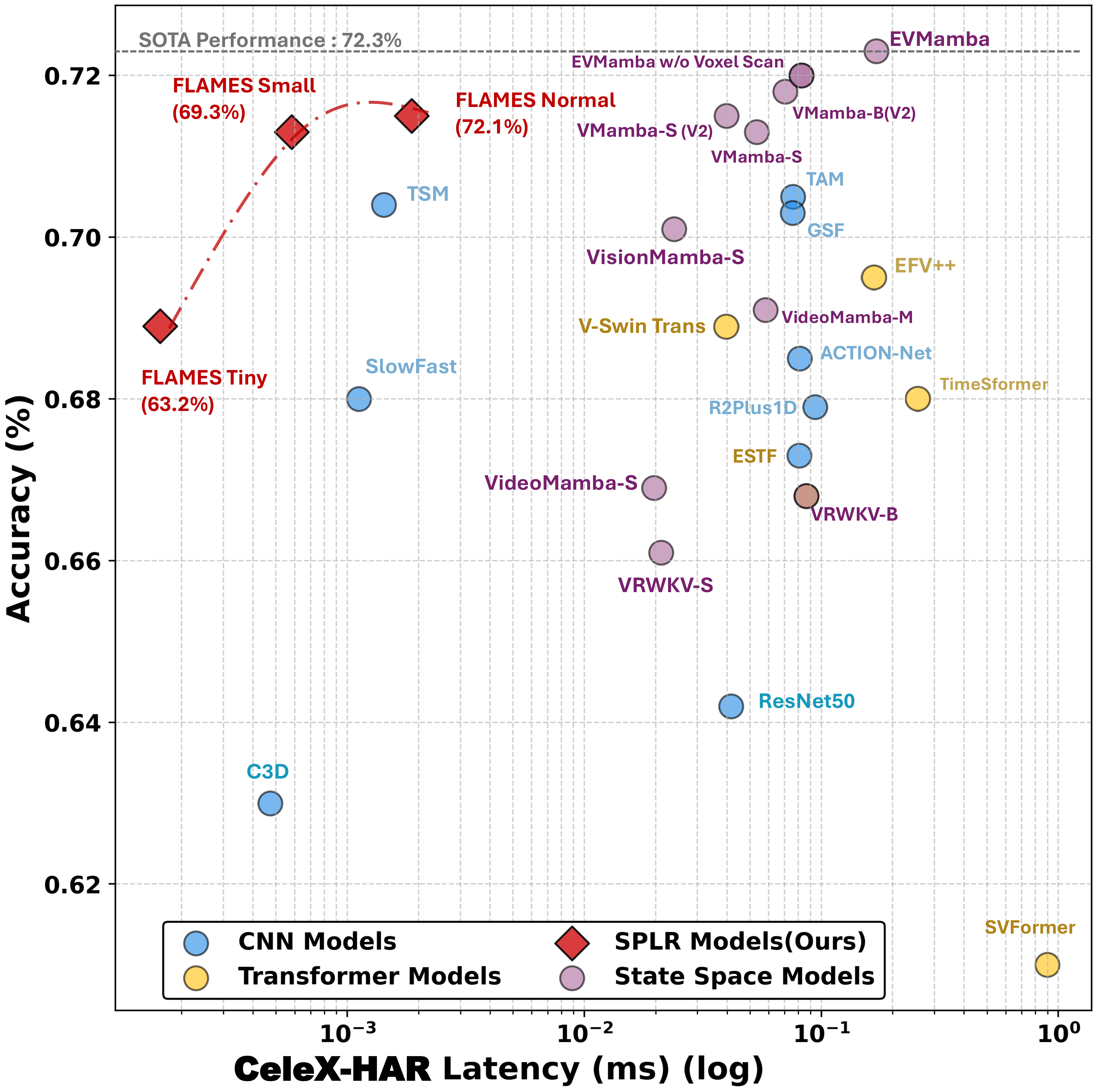}
  \end{center}
  \caption{Figure showing Accuracy vs Inference Latency for different models on the Celex-HAR dataset}
  \label{fig:latency}
  % \vspace{-5mm}
\end{figure} 

\textbf{Event Dataset Results: } 
Figure \ref{fig:flops}(a) presents the performance of our proposed FLAMES models on the DVS Gesture 128 dataset, comparing accuracy versus number of parameters with other state-of-the-art models. We evaluate three FLAMES variants—\textbf{Tiny, Small, and Normal}—with varying computational complexity and model capacity (see Table~\ref{tab:FLAMES_architecture}). FLAMES-Tiny is a lightweight model with 16 dendritic branches, 32 convolutional filters per block, and a 256-neuron readout layer, suited for low-power applications. FLAMES-Small balances efficiency and performance with 32 dendritic branches, 64 filters per block, and a 512-neuron readout. FLAMES-Normal, the full-scale model, features 64 dendritic branches, 128 filters per block, and a 1024-neuron readout for enhanced long-range memory retention. All variants process asynchronous spike inputs and employ \textbf{SA-HiPPO} for structured memory updates while maintaining efficiency in event-driven processing.

The FLAMES Normal variant achieved an accuracy of 96.5\%, effectively capturing the complex temporal dependencies in event-driven tasks. FLAMES Small and FLAMES Tiny also demonstrated competitive performance with accuracies of 93.7\% and 89.2\%, respectively, maintaining a balance between reduced parameter count and performance. Compared to other architectures like EventMamba \cite{ren2024rethinking}, TBR+I3D \cite{innocenti2021temporal}, and PointNet++ \cite{qi2017pointnetpp}, our FLAMES variants consistently showed a favorable trade-off between model complexity and accuracy. Notably, FLAMES Normal matched or even exceeded the performance of larger CNN and ViT models, such as Event Frames + I3D\cite{bi2020graph} and RG-CNN \cite{miao2019neuromorphic}, with significantly fewer parameters, emphasizing its efficiency.
We conducted an ablation study to evaluate the contribution of specific architectural components in the FLAMES models, focusing on the Dendrite Attention Layer and the SA-HiPPO matrix. Removing the dendrite mechanism led to a significant drop in accuracy across all variants, with FLAMES Normal reducing to 95.2\%. Similarly, replacing SA-HiPPO with standard LIF neurons further reduced accuracy to 90.4\%, indicating the crucial role of SA-HiPPO in maintaining long-range temporal dependencies. (Complete results are shown in Table \ref{tab:combined_comparison_dvs128_gesture} in Suppl. Sec \ref{sec:experiments}). The dashed lines in Figure \ref{fig:flops}(a) illustrate the impact of these architectural components, demonstrating the critical contribution of both Dendrite Attention and SA-HiPPO in achieving high accuracy. These results highlight the importance of each component in enabling efficient spatiotemporal learning, allowing FLAMES models to outperform other methods while maintaining fewer parameters.

We also evaluate the effectiveness of \textit{dendritic mechanisms} combined with \textit{SA-HiPPO convolutions} across SHD, SSC, and DVS Gesture datasets as detailed in Tables \ref{tab:combined_comparison_shd_ssc}, \ref{tab:combined_comparison_dvs128_gesture} (Suppl. Sec. \ref{sec:experiments}). The SSC dataset, requiring the capture of long-range temporal dependencies, proves to be more challenging than SHD and DVS Gesture.  Moreover, incorporating dendritic attention consistently enhances accuracy across all datasets, especially when using fewer channels.

\textbf{Scaling to HD Event Streams: } To evaluate the scalability of the proposed \textit{FLAMES} model, we utilized the \textit{Celex HAR} dataset\cite{wang2024event}, a high-resolution human activity recognition benchmark ($1280 \times 800$). This dataset presents significant challenges in maintaining accuracy and efficiency with large-scale spatial and temporal data. As shown in Figure \ref{fig:flops}(b), \textit{FLAMES} achieves superior accuracy compared to baseline SNNs and DNNs, maintaining high performance even at increased resolutions where other methods struggle. The \textit{SA-HiPPO convolution layer} effectively manages both spatial and temporal complexities, enabling real-time processing of HD event streams with minimal computational overhead.  Figure \ref{fig:flops}(b) also illustrates the trade-off between accuracy and computational cost (FLOPs), with \textit{FLAMES Tiny}, \textit{Small}, and \textit{Normal} achieving competitive or better accuracy compared to models like \textit{SlowFast} \cite{feichtenhofer2019slowfast} and \textit{C3D} \cite{tran2015learning}, but with significantly lower computational requirements. \textit{FLAMES Normal} exceeds the performance of models like \textit{TSM} \cite{lin2019tsm} and \textit{VisionMamba-S} \cite{zhu2024vision}, highlighting its efficiency.

\textbf{HAR-DVS: }We also evaluated on the HAR-DVS dataset \cite{wang2024hardvs} (Fig. \ref{fig:flops} c) in which our FLAMES models outperform other state-of-the-art DNN models. Unlike frame-based methods, FLAMES employs event-by-event processing to preserve temporal dynamics and introduces a novel dendritic attention mechanism, enabling efficient and robust spatio-temporal modeling. This makes FLAMES particularly well-suited for real-time event-driven applications.% [See Suppl. Sec. \ref{sec:experiments}]

\textbf{Latency Analysis: } We measure inference latency on an NVIDIA A100 (40GB VRAM) to evaluate FLAMES's efficiency for real-time applications. As shown in Figure~\ref{fig:latency}, FLAMES achieves significantly lower latency than state-of-the-art models on the Celex-HAR dataset. FLAMES-Tiny runs at 0.162 ms, the lowest among all methods, while FLAMES-Small (0.582 ms) and FLAMES-Normal (1.867 ms) maintain competitive trade-offs between latency and accuracy. In contrast, TimeSformer (255.425 ms), EFV++ (166.23 ms), and R2Plus1D (94.264 ms) exhibit much higher latencies. Even optimized models like VideoMamba-S (19.707 ms) and SlowFast (1.118 ms) are outperformed.

This demonstrates FLAMES's ability to achieve state-of-the-art efficiency with minimal computational overhead, making it ideal for real-time, resource-constrained applications. Detailed latency results are provided in Table~\ref{tab:latency_celex}.

\section{Conclusion}

In this paper we presented FLAMES - a novel hybrid paradigm that integrates the efficiency of event-driven SNNs with the structured memory of state-space models. By taking the best of both worlds, FLAMES achieves state-of-the-art performance on long-range temporal modeling while maintaining computational efficiency. This approach demonstrates that combining SNN-inspired sparse computation with continuous structured modeling can unlock new capabilities for neuromorphic computing. At its core, our approach integrates the \textbf{SA-HiPPO} mechanism, which dynamically adjusts memory retention based on inter-spike intervals, allowing efficient processing of long-range dependencies across multiple timescales. Our evaluation demonstrates state-of-the-art performance on temporal reasoning benchmarks, event-driven gesture recognition, and high-resolution neuromorphic datasets, showcasing the advantages of FLAMES in terms of efficiency, scalability, and memory retention.

\section*{Impact Statement}

This work advances energy-efficient neuromorphic computing through improved temporal processing capabilities. The primary societal impact lies in enabling more efficient AI systems that could reduce the environmental footprint of machine learning applications. Our event-driven approach may particularly benefit real-time applications in healthcare monitoring, assistive technologies, and environmental sensing, where power consumption and latency are critical constraints. However, like any advancement in AI capabilities, this technology could potentially be applied in surveillance systems or autonomous systems where ethical considerations around privacy and autonomy must be carefully considered. We encourage future development and deployment of these methods to prioritize applications that clearly benefit society while implementing appropriate safeguards against potential misuse.


\bibliography{example_paper}
\bibliographystyle{icml2025}

%%%%%%%%%%%%%%%%%%%%%%%%%%%%%%%%%%%%%%%%%%%%%%%%%%%%%%%%%%%%%%%%%%%%%%%%%%%%%%%
%%%%%%%%%%%%%%%%%%%%%%%%%%%%%%%%%%%%%%%%%%%%%%%%%%%%%%%%%%%%%%%%%%%%%%%%%%%%%%%
% APPENDIX
%%%%%%%%%%%%%%%%%%%%%%%%%%%%%%%%%%%%%%%%%%%%%%%%%%%%%%%%%%%%%%%%%%%%%%%%%%%%%%%
%%%%%%%%%%%%%%%%%%%%%%%%%%%%%%%%%%%%%%%%%%%%%%%%%%%%%%%%%%%%%%%%%%%%%%%%%%%%%%%
\newpage
\appendix
\onecolumn
% \tableofcontents

% \newpage

\section{Supplementary Section A: Detailed Proofs}
\label{sec:proofs}
% Add content here
% \section*{Theoretical Results}

\subsection{Computational Complexity of Spike-Driven SSMs}
\begin{lemma_a}
Let the spike-driven state-space model be governed by:
\[
\dot{x}(t) = A x(t) + B S(t),
\]
where \( x(t) \in \mathbb{R}^N \) is the internal state, \( A \in \mathbb{R}^{N \times N} \) is the state transition matrix, and \( S(t) \in \mathbb{R}^M \) is the input spike train. The computational complexity of updating the internal state \( x(t) \) at each spike event is \( O(N^2) \).
\end{lemma_a}

\begin{proof}
The spike-driven state-space model is governed by:
\[
\dot{x}(t) = A x(t) + B S(t),
\]
where \( x(t) \in \mathbb{R}^N \) represents the internal state of the system, \( A \in \mathbb{R}^{N \times N} \) is the state transition matrix, and \( S(t) \in \mathbb{R}^M \) represents the input spike train. When a spike event occurs at time \( t_i \), the state update can be represented by the following integral equation for \( t \in [t_i, t_{i+1}) \):
\[
x(t_i^+) = e^{A \Delta t_i} x(t_i^-) + \int_{t_i^-}^{t_i^+} e^{A (t_i^+ - \tau)} B S(\tau) \, d\tau,
\]
where:
\begin{itemize}
    \item  \( t_i^- \) and \( t_i^+ \) are the times just before and after the spike at \( t_i \),
\item \( \Delta t_i = t_i^+ - t_i^- \) is infinitesimal,
\item \( S(\tau) \) contains Dirac delta functions at spike times and is zero elsewhere.
\end{itemize}

For simplicity, we focus on the update at the spike time \( t_i \) to approximate the state transition at each event.

% %% Step 1: Matrix Exponential Computation
The update of the internal state \( x(t) \) requires computing the matrix exponential \( e^{A \Delta t} \), where \( \Delta t = t - t_i \) represents the time interval between successive spikes. Computing the exact matrix exponential for a general matrix \( A \in \mathbb{R}^{N \times N} \) is computationally expensive, involving \( O(N^3) \) operations using standard algorithms such as diagonalization or the Schur decomposition.

To reduce the computational cost, we approximate the matrix exponential using a truncated Taylor series expansion:
\[
e^{A \Delta t_i} \approx I + A \Delta t_i + \frac{1}{2} A^2 \Delta t_i^2.
\]
where \( I \) is the identity matrix of size \( N \times N \). This approximation is typically sufficient for small \( \Delta t \), which is common between spike events.

% %% Step 2: Computational Complexity of the Taylor Series Expansion
In the Taylor series expansion approximation of \( e^{A \Delta t} \), the dominant computational cost arises from multiplying the matrix \( A \in \mathbb{R}^{N \times N} \) by itself and by the state vector \( x(t) \in \mathbb{R}^N \).

% %%# 1. Computing \( A x(t) \):
The product \( A x(t) \), where \( A \in \mathbb{R}^{N \times N} \) and \( x(t) \in \mathbb{R}^N \), requires \( N^2 \) multiplications. Thus, the computational cost for this step is \( O(N^2) \).

% %%# 2. Computing \( A^2 x(t) \):
The term \( A^2 \) is computed by multiplying \( A \) by itself. Since \( A \) is an \( N \times N \) matrix, computing \( A^2 \) explicitly would have a computational cost of \( O(N^3) \). However, we avoid this by computing \( A (A x(t)) \), which involves two sequential matrix-vector products, each costing \( O(N^2) \). Therefore, the computational cost of computing \( A^2 x(t) \) is \( O(N^2) \).

% %%# 3. Computing \( B S(t) \):
The term \( B S(t) \), where \( B \in \mathbb{R}^{N \times M} \) and \( S(t) \in \mathbb{R}^M \), involves \( O(NM) \) operations. Assuming \( M \) is proportional to \( N \) or smaller, this computation contributes \( O(N^2) \) to the overall complexity.

% %%# 4. Combining the Terms:
To update the internal state \( x(t) \), we perform the following operations:
First, we multiply \( A \) by \( x(t) \): \( O(N^2) \); then multiply \( A^2 \) by \( x(t) \): \( O(N^2) \); followed by multiplying \( B \) by \( S(t) \): \( O(NM) \) and finally add the resulting vectors.

Thus, the overall computational complexity for updating the internal state \( x(t) \) at each spike event is \( O(N^2) \).

% %% Step 3: General Case and Sparsity Considerations
In the general case, where \( A \) is a dense matrix, the cost of updating the state is \( O(N^2) \). If the matrix \( A \) has a specific structure, such as being sparse or block-diagonal, the computational cost can be reduced. For example:
- If \( A \) is sparse with \( k \) non-zero entries per row, the cost of multiplying \( A \) by \( x(t) \) becomes \( O(kN) \), which can be significantly lower than \( O(N^2) \) when \( k \ll N \).
- If \( A \) is block-diagonal, the cost can be reduced to \( O(N) \) per block, depending on the number and size of the blocks. However, for the general case where no such structure is assumed, the computational complexity remains \( O(N^2) \).
The computational complexity of updating the internal state \( x(t) \) at each spike event, using the matrix exponential approximation with a Taylor series expansion, is dominated by the matrix-vector multiplication operations. Additionally, accounting for the \( B S(t) \) term maintains the overall complexity at \( O(N^2) \). Therefore, the overall computational complexity for updating the internal state at each spike event is \( O(N^2) \).

\end{proof}

\subsection{Long-Range Temporal Dependency Preservation Via Spike-Based Hippo}

\begin{theorem_a}

Let \( x(t) \in \mathbb{R}^N \) evolve according to
\[
\dot{x}(t) = A x(t) + B S(t),
\]
where:
- \( A \in \mathbb{R}^{N \times N} \) is a HiPPO matrix with all eigenvalues satisfying \( \text{Re}(\lambda_i) < 0 \) for \( i = 1, 2, \dots, N \),
- \( B \in \mathbb{R}^{N \times M} \) is the input matrix,
- \( S(t) \in \mathbb{R}^M \) is the input spike train, assumed to be bounded, i.e., there exists a constant \( S_{\infty} > 0 \) such that \( \| S(t) \| \leq S_{\infty} \) for all \( t \geq 0 \),
- \( x_0 = x(0) \in \mathbb{R}^N \) is the initial state.

Then, the spike-driven SSM preserves long-range temporal dependencies in the input spike train \( S(t) \), and the state \( x(t) \) satisfies the bound:
\[
\| x(t) \| \leq e^{-\alpha t} \| x_0 \| + \frac{\| B \| S_{\infty}}{\alpha} \left( 1 - e^{-\alpha t} \right),
\]
where \( \alpha = \min_{i} |\text{Re}(\lambda_i)| > 0 \) is the memory retention factor determined by the eigenvalues of the HiPPO matrix \( A \).

\end{theorem_a}

\begin{proof}
To establish the theorem, we will analyze the evolution of the internal state \( x(t) \) governed by the differential equation:
\[
\dot{x}(t) = A x(t) + B S(t),
\]
with initial condition \( x(0) = x_0 \).

%% Step 1: Solution via Variation of Parameters

The differential equation is a non-homogeneous linear ordinary differential equation (ODE). Using the variation of parameters method, the solution can be expressed as:
\[
x(t) = e^{A t} x_0 + \int_{0}^{t} e^{A (t - \tau)} B S(\tau) \, d\tau,
\]
where:
- \( e^{A t} x_0 \) is the solution to the homogeneous equation \( \dot{x}(t) = A x(t) \) with initial condition \( x(0) = x_0 \),
- \( \int_{0}^{t} e^{A (t - \tau)} B S(\tau) \, d\tau \) accounts for the particular solution due to the input \( S(t) \).

%% Step 2: Properties of the HiPPO Matrix \( A \)

Given that \( A \) is a HiPPO matrix, all its eigenvalues satisfy \( \text{Re}(\lambda_i) < 0 \) for \( i = 1, 2, \dots, N \). This implies that \( A \) is a Hurwitz matrix, ensuring that the system is asymptotically stable. Define the memory retention factor \( \alpha \) as:
\[
\alpha = \min_{i} |\text{Re}(\lambda_i)| > 0.
\]
This factor dictates the rate at which the influence of the initial state \( x_0 \) decays over time.

%% Step 3: Bounding the Homogeneous Solution

Consider the homogeneous solution \( e^{A t} x_0 \). Since all eigenvalues of \( A \) have negative real parts, the matrix exponential \( e^{A t} \) satisfies:
\[
\| e^{A t} \| \leq e^{-\alpha t},
\]
where \( \| \cdot \| \) denotes an operator norm (e.g., the induced 2-norm). This inequality leverages the spectral bound of \( A \) to provide an exponential decay rate.

Therefore, the contribution of the initial state is bounded by:
\[
\| e^{A t} x_0 \| \leq \| e^{A t} \| \cdot \| x_0 \| \leq e^{-\alpha t} \| x_0 \|.
\]

%% Step 4: Bounding the Particular Solution

Next, consider the particular solution:
\[
\int_{0}^{t} e^{A (t - \tau)} B S(\tau) \, d\tau.
\]
To bound its norm, apply the triangle inequality and properties of operator norms:
\[
\left\| \int_{0}^{t} e^{A (t - \tau)} B S(\tau) \, d\tau \right\| \leq \int_{0}^{t} \| e^{A (t - \tau)} \| \cdot \| B \| \cdot \| S(\tau) \| \, d\tau.
\]
Given that \( \| S(\tau) \| \leq S_{\infty} \) and \( \| e^{A (t - \tau)} \| \leq e^{-\alpha (t - \tau)} \), we have:
\[
\left\| \int_{0}^{t} e^{A (t - \tau)} B S(\tau) \, d\tau \right\| \leq \| B \| S_{\infty} \int_{0}^{t} e^{-\alpha (t - \tau)} \, d\tau.
\]
Evaluate the integral:
\[
\int_{0}^{t} e^{-\alpha (t - \tau)} \, d\tau = \int_{0}^{t} e^{-\alpha s} \, ds = \frac{1 - e^{-\alpha t}}{\alpha}.
\]
Thus, the bound becomes:
\[
\left\| \int_{0}^{t} e^{A (t - \tau)} B S(\tau) \, d\tau \right\| \leq \frac{\| B \| S_{\infty}}{\alpha} \left( 1 - e^{-\alpha t} \right).
\]

%% Step 5: Combining the Bounds

Combining the bounds for the homogeneous and particular solutions, we obtain:
\[
\| x(t) \| \leq \| e^{A t} x_0 \| + \left\| \int_{0}^{t} e^{A (t - \tau)} B S(\tau) \, d\tau \right\| \leq e^{-\alpha t} \| x_0 \| + \frac{\| B \| S_{\infty}}{\alpha} \left( 1 - e^{-\alpha t} \right).
\]
This inequality demonstrates that the influence of the initial state \( x_0 \) decays exponentially at rate \( \alpha \). Also, the accumulated influence of the input spike train \( S(t) \) is bounded and grows to a steady-state value determined by \( \| B \| \), \( S_{\infty} \), and \( \alpha \).

%% Step 6: Interpretation and Conclusion

The derived bound:
\[
\| x(t) \| \leq e^{-\alpha t} \| x_0 \| + \frac{\| B \| S_{\infty}}{\alpha} \left( 1 - e^{-\alpha t} \right),
\]
reveals that the term \( e^{-\alpha t} \| x_0 \| \) signifies that the system "forgets" its initial state exponentially fast, ensuring that old information does not dominate the state indefinitely.
Also, the integral term captures the accumulated influence of the input spike train \( S(t) \). Since \( S(t) \) is bounded, the state \( x(t) \) can retain and reflect information from the input over extended periods without being overwhelmed by the initial condition.

Therefore, the spike-driven SSM governed by a HiPPO matrix \( A \) effectively preserves long-range temporal dependencies in the input spike train \( S(t) \), while ensuring that the memory of the initial state \( x_0 \) decays at an exponential rate determined by \( \alpha \).

\end{proof}

\subsection{Error Bound For Spike-Driven Matrix Exponential Approximation}

\begin{lemma_a}
Let the matrix exponential be approximated using a Taylor expansion up to the \( n \)-th term:
\[
e^{A \Delta t} \approx I + A \Delta t + \frac{A^2 \Delta t^2}{2!} + \dots + \frac{A^n \Delta t^n}{n!}.
\]
Assume that the matrix norm \( \| \cdot \| \) is submultiplicative, i.e., \( \|AB\| \leq \|A\| \|B\| \) for any matrices \( A \) and \( B \) of compatible dimensions. Then, the error \( E_n \) of this approximation satisfies
\[
\| E_n \| \leq \frac{\| A \Delta t \|^{n+1}}{(n+1)!} e^{\| A \Delta t \|}.
\]
\end{lemma_a}

\begin{proof}
The matrix exponential can be expressed as an infinite Taylor series:
\[
e^{A \Delta t} = \sum_{k=0}^{\infty} \frac{(A \Delta t)^k}{k!}.
\]
If we truncate this series after the \( n \)-th term, the remainder \( E_n \) is given by:
\[
E_n = e^{A \Delta t} - \sum_{k=0}^{n} \frac{(A \Delta t)^k}{k!} = \sum_{k=n+1}^{\infty} \frac{(A \Delta t)^k}{k!}.
\]
To bound the norm of the error \( E_n \), we apply the submultiplicative property of the matrix norm:
\[
\| E_n \| = \left\| \sum_{k=n+1}^{\infty} \frac{(A \Delta t)^k}{k!} \right\| \leq \sum_{k=n+1}^{\infty} \frac{\| A \Delta t \|^k}{k!}.
\]
Using the submultiplicative property of the matrix norm:

\[
\| E_n \| \leq \sum_{k=n+1}^{\infty} \frac{\| A \Delta t \|^{k}}{k!}.
\]

Let \( x = \| A \Delta t \| \geq 0 \). Then:

\[
\| E_n \| \leq \sum_{k=n+1}^{\infty} \frac{x^{k}}{k!}.
\]

Since

\[
\sum_{k=n+1}^{\infty} \frac{x^{k}}{k!} = e^{x} - \sum_{k=0}^{n} \frac{x^{k}}{k!} = R_n(x),
\]

where \( R_n(x) \) is the remainder of the Taylor series expansion of \( e^{x} \).

According to Taylor's Remainder Theorem (Lagrange's form), there exists \( \xi \in [0, x] \) such that:

\[
R_n(x) = \frac{x^{n+1}}{(n+1)!} e^{\xi}.
\]

Since \( \xi \leq x \) and \( e^{\xi} \leq e^{x} \) for \( x \geq 0 \), we have:

\[
R_n(x) \leq \frac{x^{n+1}}{(n+1)!} e^{x}.
\]

Therefore:

\[
\| E_n \| \leq \frac{x^{n+1}}{(n+1)!} e^{x} = \frac{\| A \Delta t \|^{n+1}}{(n+1)!} e^{\| A \Delta t \|}.
\]

Thus, the error \( E_n \) satisfies:

\[
\| E_n \| \leq \frac{\| A \Delta t \|^{n+1}}{(n+1)!} e^{\| A \Delta t \|}.
\]
\end{proof}

% Let \( x = \| A \Delta t \| \). The series \( \sum_{k=n+1}^{\infty} \frac{x^k}{k!} \) is the tail of the exponential series \( e^{x} \). Therefore, we have:
% \[
% \sum_{k=n+1}^{\infty} \frac{x^k}{k!} = e^{x} - \sum_{k=0}^{n} \frac{x^k}{k!} \leq \frac{x^{n+1}}{(n+1)!} e^{x},
% \]
% where the inequality follows from the fact that for \( x \geq 0 \),
% \[
% \sum_{k=n+1}^{\infty} \frac{x^k}{k!} \leq \frac{x^{n+1}}{(n+1)!} e^{x}.
% \]
% This inequality can be justified by recognizing that each subsequent term in the series is bounded by \( \frac{x^{n+1}}{(n+1)!} \) multiplied by the remaining exponential growth.

% Substituting back \( x = \| A \Delta t \| \), we obtain:
% \[
% \| E_n \| \leq \frac{\| A \Delta t \|^{n+1}}{(n+1)!} e^{\| A \Delta t \|}.
% \]

---

\subsection{Boundedness Of State Trajectories In The Presence Of Spiking Inputs}

\begin{theorem_a}
\textbf{Boundedness of State Trajectory in Spike-Driven State-Space Models}

For a given initial condition \( x_0 \), the state trajectory \( x(t) \) of the FLAMES model driven by the spike input \( S(t) \) is bounded, i.e., \( \| x(t) \| \leq C \), for some constant \( C > 0 \), provided that:
\begin{enumerate}
    \item The input spikes \( S(t) \) are of finite magnitude, i.e., \( \| S(t) \| \leq S_{\infty} \) for all \( t \geq 0 \).
    \item The decay matrix \( A_S \) is Hurwitz, meaning all its eigenvalues have negative real parts.
    \item There exists a positive definite matrix \( P \) satisfying the Lyapunov equation \( A_S^T P + P A_S = -Q \), for some positive definite matrix \( Q \).
\end{enumerate}
\end{theorem_a}

\begin{proof}
Consider the FLAMES governed by:
\[
\dot{x}(t) = A_S x(t) + B S(t),
\]
where \( A_S \) is a Hurwitz matrix, \( B \) is the input matrix, and \( S(t) \) is a bounded input spike train with \( \| S(t) \| \leq S_{\infty} \) for all \( t \geq 0 \).

We define a Lyapunov function \( V(x) = x^T P x \), where \( P \) is a positive definite matrix satisfying the Lyapunov equation:
\[
A_S^T P + P A_S = -Q,
\]
with \( Q \) being a positive definite matrix. Such a \( P \) exists because \( A_S \) is Hurwitz.
The derivative of \( V(x) \) along the system trajectories is computed: 
     \[
     \dot{V}(x) = \frac{d}{dt}(x^T P x) = x^T \dot{P} x + x^T P \dot{x} + \dot{x}^T P x.
     \]
     Since \( P \) is constant (\( \dot{P} = 0 \)), and \( \dot{x} = A_S x + B S(t) \), this simplifies to:
     \[
     \dot{V}(x) = x^T P (A_S x + B S(t)) + (A_S x + B S(t))^T P x.
     \]
     Recognizing that \( P \) is symmetric (\( P^T = P \)), we can write:
     \[
     \dot{V}(x) = x^T (A_S^T P + P A_S) x + 2 x^T P B S(t).
     \]
Substituting the Lyapunov equation \( A_S^T P + P A_S = -Q \):
     \[
     \dot{V}(x) = -x^T Q x + 2 x^T P B S(t).
     \]
The term \( 2 x^T P B S(t) \) is bounded using the Cauchy-Schwarz inequality as

     \[
     2 x^T P B S(t) \leq 2 \| x \| \cdot \| P B \| \cdot \| S(t) \| \leq 2 \| P B \| S_{\infty} \| x \|.
     \]
 Next, let us define \( \gamma = 2 \| P B \| S_{\infty} \)
The derivative \( \dot{V}(x) \) becomes:

     \[
     \dot{V}(x) \leq - x^T Q x + \gamma \| x \|.
     \]
Since \( Q \) is positive definite, \( x^T Q x \geq \lambda_{\text{min}}(Q) \| x \|^2 \), where \( \lambda_{\text{min}}(Q) \) is the smallest eigenvalue of \( Q \). Therefore:

     \[
     \dot{V}(x) \leq - \lambda_{\text{min}}(Q) \| x \|^2 + \gamma \| x \|.
     \]
Completing the square:
     \[
     \dot{V}(x) \leq - \lambda_{\text{min}}(Q) \left( \| x \|^2 - \frac{\gamma}{\lambda_{\text{min}}(Q)} \| x \| \right) = - \lambda_{\text{min}}(Q) \left( \| x \| - \frac{\gamma}{2 \lambda_{\text{min}}(Q)} \right)^2 + \frac{\gamma^2}{4 \lambda_{\text{min}}(Q)}.
     \]
This inequality indicates that \( \dot{V}(x) < 0 \) whenever \( \| x \| > \frac{\gamma}{2 \lambda_{\text{min}}(Q)} \).  Since \( V(x) \geq 0 \) and \( \dot{V}(x) \) is negative outside a ball of radius \( C = \frac{\gamma}{2 \lambda_{\text{min}}(Q)} \), the state \( x(t) \) will ultimately remain within this bounded region. Therefore, \( \| x(t) \| \leq C \) for all \( t \geq 0 \)
 
\end{proof}

\newpage
\section{Supplementary Section B: Extended Experimental Results}
\label{sec:experiments}
\subsection{Datasets and Tasks}

\hlt{
In this study, we evaluate the performance of the FLAMES model across a diverse set of datasets, each presenting unique challenges in event-driven processing. The datasets include Sequential CIFAR-10, Sequential CIFAR-100 \cite{krizhevsky2009learning}, DVS Gesture \cite{amir2017low}, HAR-DVS \cite{wang2024hardvs}, Celex-HAR \cite{wang2024event}, Long Range Arena (LRA) \cite{tay2020long}, Spiking Heidelberg Digits (SHD) \cite{cramer2020heidelberg}, and Spiking Speech Commands (SSC). For all experiments, the FLAMES model processes inputs on an event-by-event basis, leveraging its temporal dynamics to handle fine-grained temporal dependencies without accumulating events into frames. Below, we provide detailed descriptions of each dataset and the corresponding experimental setups.

\textbf{Sequential CIFAR-10 and CIFAR-100}: The CIFAR-10 and CIFAR-100 datasets \cite{krizhevsky2009learning} consist of \(32 \times 32\) RGB images across 10 and 100 classes, respectively. To simulate a temporal sequence, each image is divided into 16 non-overlapping patches of size \(8 \times 8\) pixels. These patches are presented to the model sequentially in a raster-scan order, from top-left to bottom-right. Each patch is treated as an independent event in the sequence. The task involves classifying the image based on the full sequence of patches, requiring the model to integrate information over the entire sequence. This setup evaluates the model's ability to process spatial information in a temporal context.

\textbf{DVS Gesture Dataset}: The DVS Gesture dataset \cite{amir2017low} comprises recordings from a Dynamic Vision Sensor (DVS), capturing 11 hand gestures performed by 29 subjects under varying lighting conditions. Each event is characterized by its spatial location \((x, y)\), timestamp \(t\), and polarity \(p\) (on/off). The dataset provides a challenging benchmark for models to recognize dynamic gestures from sparse, asynchronous event streams. In our experiments, the FLAMES model processes each event individually as it occurs, without accumulating them into temporal frames, thereby maintaining high temporal resolution and reducing latency.

\textbf{HAR-DVS Dataset}: The HAR-DVS dataset \cite{wang2024hardvs} contains neuromorphic event streams representing human activities, recorded with a DVS. Activities include walking, running, and other movement-based tasks. Each event is defined by its spatial coordinates, timestamp, and polarity. The dataset tests the model's ability to recognize complex human activities from sparse event streams. The FLAMES model processes each spike event-by-event, dynamically updating its internal state for each incoming spike, enabling precise temporal modeling of the activity sequences.

\textbf{Celex-HAR Dataset}: The Celex-HAR dataset \cite{wang2024event} consists of high-resolution event streams captured with a CeleX camera for human activity recognition. Activities include actions such as sitting, standing, and walking. Each event is represented by its spatial coordinates, timestamps, and polarity. The dataset provides a comprehensive benchmark for evaluating models on high-resolution event-based data. The FLAMES model processes each spike event-by-event, allowing it to capture the fine-grained temporal dynamics of human activities.

\textbf{Long Range Arena (LRA)}: The Long Range Arena benchmark \cite{tay2020long} evaluates a model's ability to process long sequences and capture dependencies over extended temporal horizons. Tasks such as ListOps and Path-X involve sequence lengths ranging from hundreds to thousands of tokens. Although these tasks involve discrete tokens rather than spikes, we simulate event-driven processing by treating each token as an individual event presented sequentially. The FLAMES model leverages its temporal dynamics to capture long-range dependencies efficiently.

\textbf{Spiking Heidelberg Digits (SHD) and Spiking Speech Commands (SSC)}: The SHD and SSC datasets \cite{cramer2020heidelberg} are benchmarks for spiking neural networks, containing neuromorphic spike streams derived from speech datasets. SHD consists of spoken digit recordings converted to spike trains using the CochleaAMS model, while SSC contains spiking representations of spoken command audio, representing keywords like "yes," "no," and "stop." Each event is characterized by its spatial location, timestamp, and polarity. The datasets evaluate the model's performance on tasks involving complex spatio-temporal patterns in speech data. The FLAMES model processes each spike event as it occurs, dynamically updating its state, ensuring high temporal resolution and efficient processing for speech recognition tasks.

\par Across all datasets, the FLAMES model processes inputs on an event-by-event basis. This approach allows it to maintain high temporal resolution and capture fine-grained spatio-temporal patterns, distinguishing it from frame-based methods. The event-by-event design also reduces computational overhead and ensures low latency, making the model well-suited for real-time applications.

}

\begin{table}[h!]
\centering
\caption{\hlt{Comparison of FLAMES models with state-of-the-art on the HARDVS dataset. Accuracy is measured in percentage, and computational cost is in GFLOPs.}}
\label{tab:FLAMES_comparison}
\hlt{
\begin{tabular}{lcc}
\toprule
\textbf{Model}                 & \textbf{GFLOPs} & \textbf{Accuracy (\%)} \\
\midrule
C3D \cite{tran2015learning}    & 0.1             & 50.52                 \\
R2Plus1D \cite{tran2018closer} & 20.3            & 49.06                 \\
TSM \cite{lin2019tsm}          & 0.3             & 52.63                 \\
ACTION-Net \cite{wang2021action} & 17.3          & 46.85                 \\
TAM \cite{liu2021tam}          & 16.6            & 50.41                 \\
V-SwinTrans \cite{liu2022video} & 8.7            & 51.91                 \\
SlowFast \cite{feichtenhofer2019slowfast} & 0.3  & 46.54                 \\
ESTF \cite{wang2024hardvs}     & 17.6            & 51.22                 \\
ExACT \cite{zhou2024exact}     & 1.3             & 90.10                 \\
% SpikMamba  \cite{chen2024spikmamba}            & 0.12            & 97.32        \\
\midrule
\textbf{FLAMES-Tiny  \textit{[Ours]} }                   & \textbf{0.034}           & \textbf{65.42 }                 \\
\textbf{FLAMES-Small  \textit{[Ours]} }                   & \textbf{0.13}            & \textbf{79.36}                  \\
\textbf{FLAMES-Normal   \textit{[Ours]}  }                &\textbf{ 0.41 }           & \textbf{88.29 }                \\
\bottomrule
\end{tabular}
}
\end{table}

%65.42, 79.36, 88.29

\begin{table}[h]
    \centering
    \caption{Detailed Architecture of FLAMES Models (Tiny, Small, and Normal)}
    \label{tab:FLAMES_architecture}
     \resizebox{\columnwidth}{!}{%
    \begin{tabular}{|c|c|c|c|}
        \hline
        \textbf{Layer Type} & \textbf{FLAMES Tiny} & \textbf{FLAMES Small} & \textbf{FLAMES Normal} \\ \hline
        \textbf{Input Representation} & \multicolumn{3}{c|}{Asynchronous Spike Events $(x, y, t, p)$} \\ \hline
        
        \multirow{2}{*}{\textbf{Dendrite Attention Layer}} & 16 dendritic branches & 32 dendritic branches & 64 dendritic branches \\
        & $\tau_d = [\tau_1, \ldots, \tau_{16}]$ & $\tau_d = [\tau_1, \ldots, \tau_{32}]$ & $\tau_d = [\tau_1, \ldots, \tau_{64}]$ \\ \hline
        
        \multirow{2}{*}{\textbf{Convolutional Block 1}} & Conv2D (32 filters, 3x3) & Conv2D (64 filters, 3x3) & Conv2D (128 filters, 3x3) \\ 
        & Batch Norm, Max Pool (2x2) & Batch Norm, Max Pool (2x2) & Batch Norm, Max Pool (2x2) \\ \hline

        \multirow{2}{*}{\textbf{Convolutional Block 2}} & Conv2D (32 filters, 3x3) & Conv2D (64 filters, 3x3) & Conv2D (128 filters, 3x3) \\ 
        & Batch Norm, Max Pool (2x2) & Batch Norm, Max Pool (2x2) & Batch Norm, Max Pool (2x2) \\ \hline

        \textbf{Spatial Pooling Layer} & Pool (2x2) & Pool (2x2) & Pool (2x2) \\ \hline
        
        \textbf{FLAMES Convolution} & \multicolumn{3}{c|}{State Update using Spike-Aware HiPPO and NPLR decomposition for efficient event-driven convolution} \\ \hline
        
        \multirow{2}{*}{\textbf{Normalization Layer}} & Layer Norm & Layer Norm & Layer Norm \\ 
        & \multicolumn{3}{c|}{Normalizes the state variables to stabilize training} \\ \hline

        \multirow{2}{*}{\textbf{Readout Layer}} & Fully Connected (256 neurons) & Fully Connected (512 neurons) & Fully Connected (1024 neurons) \\
        & Softmax for classification & Softmax for classification & Softmax for classification \\ \hline
    \end{tabular}
    }
\end{table}

\begin{figure}
    \centering
    \includegraphics[width=\linewidth]{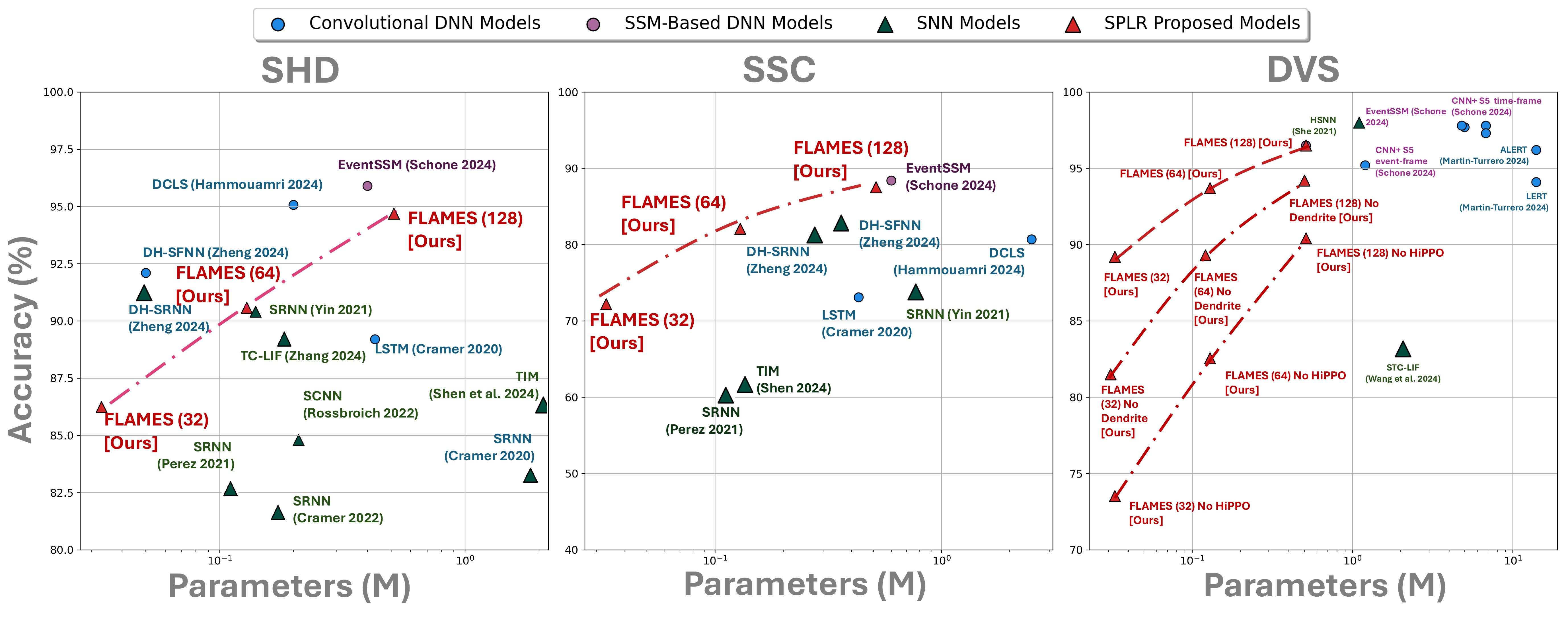}
    \caption{\hlt{Comparison of our FLAMES to the state-of-the-art on DVS128-Gesture\cite{amir2017low}, Spiking Heidelberg digits (SHD) and Spiking Speech Commands (SSC) \cite{cramer2020heidelberg}  datasets}}
    \label{fig:shd_ssc}
\end{figure}

\begin{figure}
    \centering
    \includegraphics[width=0.7\linewidth]{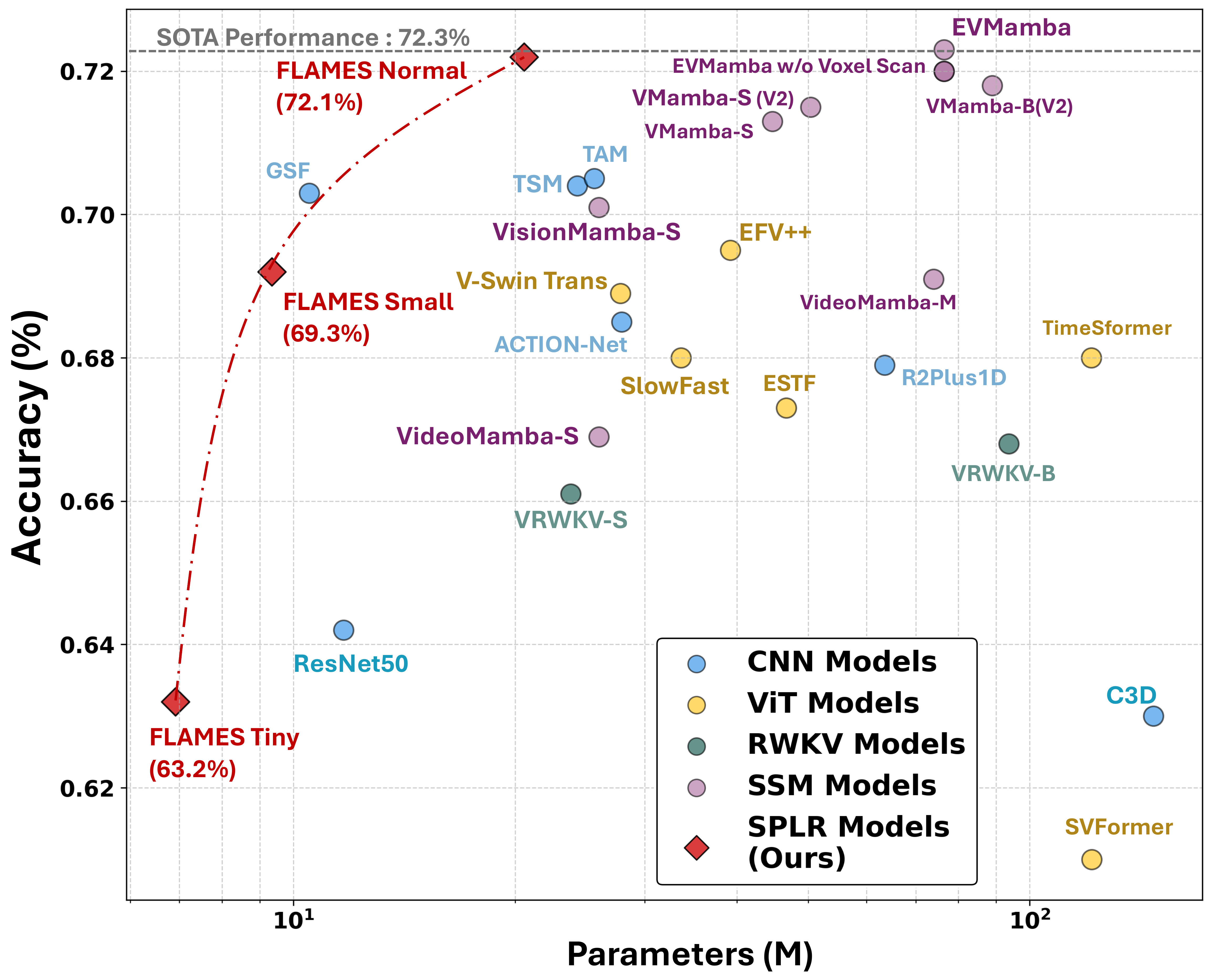}
    \caption{Figure showing the Parameters vs Accuracy of different state of the art DNN and SNN models on the Celex-HAR \cite{wang2024event} dataset wrt the FLAMES models}
    \label{fig:celex_param_vs_acc}
\end{figure}

\begin{table*}[!htp]
\small      
\center 
\caption{Experimental results on CeleX-HAR dataset.~}  
\label{CeleX_CeleXResults}
\begin{tabular}{c|l|c|c|c|c|c|c} 
\hline \toprule [0.5 pt] 
\textbf{No.} &\textbf{Algorithm}  &\textbf{Publish}  &\textbf{Arch.} &\textbf{FLOPs}  &\textbf{Params}  &\textbf{acc/top-1}  &\textbf{Code} \\ 
\hline 
01 & ResNet-50~\cite{he2016deep}  &CVPR-2016    &CNN  &8.6G  &11.7M  &0.642   &\href{https://github.com/KaimingHe/deep-residual-networks}{URL}      \\	
02 & ConvLSTM~\cite{shi2015convolutional}  &NIPS-2015     &CNN, LSTM   &-  &- &0.539   &\href{https://github.com/ndrplz/ConvLSTM_pytorch}{URL}      \\		
03 & C3D~\cite{tran2015learning}      &ICCV-2015      &CNN  &0.1G &147.2M &0.630  &\href{https://github.com/leftthomas/R2Plus1D-C3D}{URL}       \\	
04 & R2Plus1D~\cite{tran2018closer}  &CVPR-2018     &CNN  &20.3G &63.5M &0.679    &\href{https://github.com/leftthomas/R2Plus1D-C3D}{URL}       \\
05 & TSM~\cite{lin2019tsm}      &ICCV-2019     &CNN  &0.3G &24.3M  &0.704  &\href{https://github.com/mit-han-lab/temporal-shift-module}{URL}      \\	
06 & ACTION-Net~\cite{wang2021action}  &CVPR-2021      &CNN &17.3G &27.9M  &0.685   &\href{https://github.com/V-Sense/ACTION-Net}{URL}      \\
07 & TAM~\cite{liu2021tam} 	  &ICCV-2021        &CNN  &16.6G &25.6M  &0.705   &\href{https://github.com/liu-zhy/temporal-adaptive-module}{URL}       \\
08 & GSF~\cite{sudhakaran2023gate} &TPAMI-2023  &CNN   &16.5G     &10.5M       & 0.703         &\href{https://github.com/swathikirans/GSF}{URL}  \\
% 09 & Wang et al.~\cite{wang2024generative}  &AAAI-2024  &CNN &    &        &       &      &\href{https://github.com/aaai-24/Generative-based-KD}{URL}  \\
\hline 
09 & V-SwinTrans~\cite{liu2022video} &CVPR-2022      &ViT  &8.7G &27.8M  &0.689  &\href{https://github.com/SwinTransformer/Video-Swin-Transformer}{URL}       \\
10 & TimeSformer~\cite{bertasius2021space}   &ICML-2021       &ViT   &53.6G &121.2M &0.680  &\href{https://github.com/facebookresearch/TimeSformer}{URL}       \\
11 & SlowFast~\cite{feichtenhofer2019slowfast} &ICCV-2019      &ViT   &0.3G &33.6M &0.680   &\href{https://github.com/facebookresearch/SlowFast}{URL}      \\
12 & SVFormer~\cite{xing2023svformer}    &CVPR-2023     &ViT   &196.0G   &121.3M   &0.610    &\href{https://github.com/ChenHsing/SVFormer}{URL}  \\
13 & EFV++~\cite{chen2024retain}   &arXiv-2024      &ViT, GNN   &36.3G   &39.2M    &0.695  &\href{https://github.com/Event-AHU/EFV_event_classification/tree/EFVpp}{URL}      \\
14 & ESTF~\cite{wang2024hardvs}  &AAAI-2024  &ViT, CNN   & 17.6G       & 46.7M   &0.673       &\href{https://github.com/Event-AHU/HARDVS}{URL}  \\
% 16 & OST~\cite{chen2024ost} &CVPR-2024  &ViT       &           &           &         &                &\href{https://github.com/tomchen-ctj/OST}{URL} \\
\hline 
15 & VRWKV-S~\cite{duan2024vrwkv}  &arXiv-2024  &RWKV  &4.6G   &23.8M    &0.661   &\href{https://github.com/OpenGVLab/Vision-RWKV}{URL}  \\
16 & VRWKV-B~\cite{duan2024vrwkv}  &arXiv-2024  &RWKV  &18.2G   &93.7M   &0.668  &\href{https://github.com/OpenGVLab/Vision-RWKV}{URL}   \\
\hline
17    &Vision Mamba-S~\cite{zhu2024vision}  &ICML-2024       &SSM   &5.1G   &26.0M    &0.701&\href{https://github.com/hustvl/Vim}{URL}     \\
18    &VMamba-S~\cite{liu2024VMamba}  &arXiv-2024       &SSM   &11.2G   &44.7M    &0.713   &\href{https://github.com/MzeroMiko/VMamba}{URL}     \\
19    &VMamba-S(V2)~\cite{liu2024VMamba}  &arXiv-2024       &SSM   &8.7G   &50.4M    &0.715   &\href{https://github.com/MzeroMiko/VMamba}{URL}      \\
20    &VMamba-B~\cite{liu2024VMamba}   &arXiv-2024      &SSM   &18.0G   &76.5M    &0.720 &\href{https://github.com/MzeroMiko/VMamba}{URL}       \\
21    &VMamba-B(V2)~\cite{liu2024VMamba}  &arXiv-2024       &SSM   &15.4G   &88.9M    &0.718  &\href{https://github.com/MzeroMiko/VMamba}{URL}      \\
22    &VideoMamba-S~\cite{li2024videomamba}   &ECCV-2024   &SSM  &4.3G  &26.0M    &0.669 &\href{https://github.com/OpenGVLab/VideoMamba}{URL}      \\
23   &VideoMamba-M~\cite{li2024videomamba}   &ECCV-2024   &SSM  &12.7G  &74.0M  &0.691 &\href{https://github.com/OpenGVLab/VideoMamba}{URL}      \\

24   & EVMamba   & arXiv-2024      &SSM   &37.2G             &76.5M    &\textbf{0.723}     & \href{https://github.com/Event-AHU/CeleX-HAR/tree/main}{URL}    \\
25   & EVMamba $w/o$ Voxel Scan   & arXiv-2024     &SSM   &18.0G   &76.5M    &0.720   & \href{https://github.com/Event-AHU/CeleX-HAR/tree/main}{URL}    \\ \hline
26   &\textbf{FLAMES-Tiny (Ours)}   &-      &SSM   &0.034G   & 7.91M    &0.632   &-     \\
27   &\textbf{FLAMES-Small (Ours)}   &-      &SSM   &0.13G   &13.35M    &0.692   &-     \\
28   &\textbf{FLAMES-Normal (Ours)}   &-      &SSM   &0.41G   &25.57M    &0.722   &-     \\
\hline \toprule [0.5 pt]  
\end{tabular}
\end{table*}

\begin{table}[ht]
\centering
\caption{Comparison of classification accuracy and parameters of different models across SHD and SSC datasets.}
\label{tab:combined_comparison_shd_ssc}
\resizebox{\textwidth}{!}{%
\hlt{
\begin{tabular}{lcccc}
\hline
\textbf{Model} & \multicolumn{2}{c}{\textbf{SHD}} & \multicolumn{2}{c}{\textbf{SSC}} \\ 
               & \textbf{\#Parameters} & \textbf{Accuracy (\%)} & \textbf{\#Parameters} & \textbf{Accuracy (\%)} \\ \hline
SFNN \cite{cramer2020heidelberg} & 0.09 M & 48.1 & 0.09 M & 32.5 \\
SRNN \cite{cramer2020heidelberg} & 1.79 M & 83.2 & - & - \\
SRNN \cite{cramer2022surrogate} & 0.17 M & 81.6 & - & - \\
SRNN \cite{perez2021neural} & 0.11 M & 82.7 & 0.11 M & 60.1 \\
SCNN \cite{rossbroich2022fluctuation} & 0.21 M & 84.8 & - & - \\
SRNN \cite{yin2021accurate} & 0.14 M & 90.4 & 0.77 M & 74.2 \\
HRSNN \cite{chakraborty2023heterogeneous_iclr23} & - & 80.01 & - & 59.28 \\
LSTM \cite{cramer2020heidelberg} & 0.43 M & 89.2 & 0.43 M & 73.1 \\
DH-SRNN \cite{zheng2024temporal} & 0.05 M & 91.34 & 0.27 M & 81.03 \\
DH-SFNN \cite{zheng2024temporal} & 0.05 M & 92.1 & 0.35 M & 82.46 \\
ASGL \cite{wang2023adaptive} & - & 78.90 & - & 78.90 \\
DCLS \cite{hammouamri2024learning} & 0.2 M & 95.07 & 2.5 M & 80.69 \\
TIM \cite{shen2024tim} & 2.59 M & 86.3 & 0.111 M & 61.09 \\
TC-LIF \cite{zhang2024tc} & 0.142 M & 88.91 & - & - \\
\hline
FLAMES-Normal (128) \textbf{[Ours]} & 0.513 M & 94.68 & 0.513 M & 87.52  \\
FLAMES-Small (64) \textbf{[Ours]} & 0.129 M & 90.57 & 0.129 M & 82.08  \\
FLAMES-Tiny (32) \textbf{[Ours]} & 0.033 M & 86.24 & 0.033 M & 72.19 \\ \hline
\end{tabular}%
}
}
\end{table}

\begin{table}[ht]
\centering
\caption{Comparison of classification accuracy, parameters, and FLOPs of different models across the DVS128-Gesture dataset.}
\label{tab:combined_comparison_dvs128_gesture}
\resizebox{\textwidth}{!}{%
\hlt{
\begin{tabular}{lccc}
\hline
\textbf{Model} & \textbf{\#Parameters (M)} & \textbf{GFLOPs} & \textbf{Accuracy (\%)} \\ \hline
Yousefzadeh et al. \cite{yousefzadeh2019asynchronous} & 1.2 & - & 95.2 \\
Xiao et al. \cite{xiao2022online} & - & - & 96.9 \\
RTRL \cite{subramoney2023efficient} & 4.8 & - & 97.8 \\
She et al. \cite{she2021sequence} & 1.1 & - & 98.0 \\
Liu et al. \cite{liu2022fast} & - & - & 98.8 \\
Chakraborty et al. \cite{chakraborty2023heterogeneous_frontiers} & - & - & 96.5 \\
Martin-Turrero et al. \cite{martin2024alert} & 14 & - & 96.2 \\
Martin-Turrero et al. \cite{martin2024alert} & 14 & - & 94.1 \\
CNN + S5 (time-frames) \cite{schone2024scalable} & 6.8 & - & 97.8 \\
Event-SSM \cite{schone2024scalable} & 5 & - & 97.7 \\
CNN + S5 (event-frames) \cite{schone2024scalable} & 6.8 & - & 97.3 \\
TBR+I3D \cite{innocenti2021temporal} & 12.25 & 38.82 & 99.6 \\
Event Frames + I3D \cite{bi2020graph} & 12.37 & 30.11 & 96.5 \\
EV-VGCNN \cite{deng2022voxel} & 0.82 & 0.46 & 95.7 \\
RG-CNN \cite{miao2019neuromorphic} & 19.46 & 0.79 & 96.1 \\
PointNet++ \cite{wang2019space} & 1.48 & 0.872 & 95.3 \\
PLIF \cite{fang2021incorporating} & 1.7 & - & 97.6 \\
GET \cite{peng2023get} & 4.5 & - & 97.9 \\
Swin-T v2 \cite{liu2022swin} & 7.1 & - & 93.2 \\
TTPOINT \cite{ren2024rethinking} & 0.334 & 0.587 & 98.8 \\
EventMamba \cite{ren2024rethinking} & 0.29 & 0.219 & 99.2 \\ 
STC-LIF \cite{zuo2024temporal} & 3.922 & - & 83.0 \\
Spike-Driven Transformer \cite{yao2024spike} & 36.01 & 33.32 & 99.3 \\
\hline
FLAMES-Normal (128) \textbf{[Ours]} & 0.513 & 0.43 & 96.5 \\
FLAMES-Small (64) \textbf{[Ours]} & 0.129 & 0.14 & 93.7 \\
FLAMES-Tiny (32) \textbf{[Ours]} & 0.033 & 0.07 & 89.2 \\ \hline
FLAMES-Normal (128 Channels) No Dendrite \textbf{[Ours - Ablation]} & 0.501 & 0.43 & 95.2 \\
FLAMES-Small (64 Channels) No Dendrite \textbf{[Ours - Ablation]} & 0.121 & 0.14 & 89.3 \\
FLAMES-Tiny (32 Channels) No Dendrite \textbf{[Ours - Ablation]} & 0.031 & 0.07 & 81.5 \\ \hline
FLAMES-Normal (128 Channels) No HiPPO \textbf{[Ours - Ablation]} & 0.501 & 0.43 & 90.4 \\
FLAMES-Small (64 Channels) No HiPPO \textbf{[Ours - Ablation]} & 0.121 & 0.14 & 82.6 \\
FLAMES-Tiny (32 Channels) No HiPPO \textbf{[Ours - Ablation]} & 0.031 & 0.07 & 73.5 \\
\hline
\end{tabular}%
}
}
\end{table}

\hlt{
\subsection{Ablation Studies: } 
To evaluate the contribution of individual components in the FLAMES model, we performed extensive ablation studies on the sequential CIFAR-10 dataset. Specifically, we analyzed the impact of removing or replacing key components such as the dendritic attention layer, Spike-Aware HiPPO (SA-HiPPO), NPLR decomposition, and FFT convolution. The results of these experiments, along with the corresponding model parameters and computational costs (in GFLOPs), are summarized in Table~\ref{tab:ablation_FLAMES_seqcifar10_flops}.

\textbf{Impact of Dendritic Attention Layer}
Removing the dendritic attention mechanism leads to a reduction in both accuracy and model parameters. The accuracy drops across all channel configurations, with the largest channels (128) seeing a decrease from 90.25\% to 85.83\%. The smaller channel configurations (64 and 32) experience similar drops, highlighting the dendritic attention's role in improving the spatio-temporal feature representation. Interestingly, removing this mechanism slightly reduces the model's GFLOPs since the computations associated with the dendritic layer are avoided.

\textbf{Impact of Spike-Aware HiPPO}
Replacing SA-HiPPO with a simple LIF-based mechanism leads to a moderate drop in accuracy (e.g., from 90.25\% to 87.62\% for 128 channels). However, this modification does not alter the computational cost (GFLOPs), as SA-HiPPO primarily affects the temporal memory adaptation rather than the core matrix or convolution operations. These results emphasize SA-HiPPO's critical role in retaining and managing temporal dynamics effectively.

\textbf{Impact of NPLR Decomposition}
The NPLR decomposition significantly reduces the computational complexity of state-space updates. Removing NPLR decomposition results in a notable increase in GFLOPs across all configurations (e.g., from 0.43 GFLOPs to 1.8 GFLOPs for 128 channels) due to the quadratic complexity of dense matrix operations. Despite this computational overhead, the accuracy remains relatively stable, highlighting that NPLR's primary advantage is computational efficiency rather than feature extraction performance.

\textbf{Impact of FFT Convolution}
FFT convolution is integral to efficiently handling long-range temporal dependencies. Replacing FFT convolution with standard time-domain convolution increases the GFLOPs substantially (e.g., from 0.43 GFLOPs to 1.2 GFLOPs for 128 channels). Furthermore, the accuracy sees a more pronounced decline (e.g., from 90.25\% to 86.47\%), particularly in tasks requiring high temporal resolution. These results underscore FFT convolution's dual role in reducing computational cost and maintaining temporal modeling performance.

\textbf{Summary of Findings}
The ablation studies validate the critical importance of each component in the FLAMES model:
\begin{itemize}
    \item The dendritic attention layer enhances the spatio-temporal feature representation, significantly improving accuracy.
    \item SA-HiPPO dynamically adjusts temporal memory retention, contributing to performance robustness without additional computational overhead.
    \item NPLR decomposition ensures scalability by reducing the computational cost of state-space updates, making the model efficient for large-scale tasks.
    \item FFT convolution is indispensable for capturing long-range dependencies efficiently while keeping computational complexity low.
\end{itemize}
The full FLAMES model represents a carefully optimized design that balances accuracy, efficiency, and scalability, making it suitable for real-time and resource-constrained spiking neural network applications.
}

\subsection{Long-Range Dependencies}

\begin{table}[h]
\centering
\caption{Updated Ablation Study for FLAMES Variants on seqCIFAR-10 with FLOPs}
\label{tab:ablation_FLAMES_seqcifar10_flops}
\resizebox{\columnwidth}{!}{%
\hlt{
\begin{tabular}{ccccc}
\hline
\textbf{Model Variant}             & \textbf{Channels} & \textbf{Accuracy (\%)} & \textbf{Params (M)} & \textbf{FLOPs (GFLOPs)} \\ \hline
FLAMES (Full)                        & 128               & 90.25                  & 0.513               & 0.43                     \\ 
FLAMES (No SA-HiPPO)                 & 128               & 87.62                  & 0.501               & 0.43                     \\ 
FLAMES (No NPLR Decomposition)       & 128               & 88.05                  & 0.513               & 1.8                      \\ 
FLAMES (No FFT Convolution)          & 128               & 86.47                  & 0.513               & 1.2                      \\ 
FLAMES (No Dendrite)                 & 128               & 85.83                  & 0.501               & 0.43                     \\ \hline
FLAMES (Full)                        & 64                & 88.62                  & 0.129               & 0.14                     \\ 
FLAMES (No SA-HiPPO)                 & 64                & 86.14                  & 0.121               & 0.14                     \\ 
FLAMES (No NPLR Decomposition)       & 64                & 86.72                  & 0.129               & 0.56                     \\ 
FLAMES (No FFT Convolution)          & 64                & 85.23                  & 0.129               & 0.32                     \\ 
FLAMES (No Dendrite)                 & 64                & 84.65                  & 0.121               & 0.14                     \\ \hline
FLAMES (Full)                        & 32                & 83.15                  & 0.033               & 0.034                    \\ 
FLAMES (No SA-HiPPO)                 & 32                & 81.75                  & 0.031               & 0.034                    \\ 
FLAMES (No NPLR Decomposition)       & 32                & 82.12                  & 0.033               & 0.12                     \\ 
FLAMES (No FFT Convolution)          & 32                & 80.62                  & 0.033               & 0.08                     \\ 
FLAMES (No Dendrite)                 & 32                & 80.05                  & 0.031               & 0.034                    \\ \hline
\end{tabular}%
}
}
\end{table}

\textbf{Sequential CIFAR Datasets}
The first set of experiments evaluates the ability of the proposed \textit{FLAMES} model to effectively capture long-range dependencies in sequential data. This is crucial for applications involving event-driven data spanning extended periods, such as continuous gesture recognition and video analysis. To simulate long-term temporal relationships, we conduct experiments using the \textit{Sequential CIFAR-10} and \textit{Sequential CIFAR-100} datasets, where each image is transformed into a sequence of frames.

In these experiments, we compare the performance of \textit{FLAMES} against several baselines, including traditional SNN models. The key focus is on assessing the effectiveness of our \textit{Spike-Aware HiPPO (SA-HiPPO)} dynamics in retaining temporal memory over long sequences. The results are presented in Table \ref{tab:sequential}, which includes classification accuracy for different sequence lengths, as well as model complexity in terms of the number of parameters.

As seen in Table \ref{tab:sequential}, \textit{FLAMES} significantly outperforms the baselines in capturing long-range dependencies. The \textit{FLAMES} model with 128 channels achieves an accuracy of 90.25\% on the \textit{Sequential CIFAR-10} dataset and 65.33\% on \textit{Sequential CIFAR-100}, which surpasses the performance of all baseline models by a substantial margin. These results indicate that \textit{FLAMES} not only maintains memory over extended input sequences but also converges faster, achieving higher accuracy with fewer epochs compared to traditional spiking and hybrid models.

The ablation study further reveals that the SA-HiPPO matrix incorporated in \textit{FLAMES} plays a pivotal role in enhancing temporal filtering capabilities, leading to improved convergence rates and more robust performance in long-range dependency tasks. This improvement is evident in the accuracy gains observed in \textit{FLAMES} compared to other models, including those using mechanisms like GLIF and PLIF.

Moreover, even when model complexity is reduced, as seen in the \textit{FLAMES} variants with 64 and 32 channels, our model maintains superior accuracy compared to all baseline architectures. For instance, the \textit{FLAMES} with 64 channels achieves 88.\% accuracy on \textit{Sequential CIFAR-10}, outperforming other models with similar parameter counts, demonstrating the efficiency and scalability of the proposed SA-HiPPO dynamics for capturing long-term dependencies in sequential data.

These findings validate the superior temporal modeling capabilities of \textit{FLAMES}, making it well-suited for tasks that require efficient and scalable handling of long-range dependencies in sequential, event-driven data.

\textbf{Long Range Arena Datasets:} We evaluate the ability of the proposed \textit{FLAMES} model to capture long-range dependencies using the \textbf{Long Range Arena (LRA)} dataset \cite{tay2020long}. The LRA benchmark evaluates models on tasks requiring long-context understanding, where Transformer-based non-spiking models often exhibit suboptimal performance due to the computational overhead of attention mechanisms, which scales poorly with increasing sequence lengths. As shown in Table \ref{tab:lra}, we benchmark our method against state-of-the-art alternatives, including the LMU-based spiking model, SpikingLMUFormer \cite{liu2024lmuformer}, and the BinaryS4D model \cite{stan2024learning}. While BinaryS4D is not fully spiking—it relies on floating-point MAC operations for matrix multiplications—it incorporates LIF neurons to spike from an underlying state-space model (SSM), providing a hybrid approach to handling long-range dependencies.

\begin{table}[h]
\centering
\caption{Comparison of Architectures on Sequential CIFAR-10 and CIFAR-100}
\label{tab:sequential}
\resizebox{0.7\columnwidth}{!}{%
\begin{tabular}{cccccc}
\hline
\multirow{2}{*}{\textbf{Architecture}} & \multirow{2}{*}{\textbf{Channels}} & \multirow{2}{*}{\textbf{Layer Type}} & \multicolumn{1}{c}{\textbf{seqCIFAR10}} & \multicolumn{1}{c}{\textbf{seqCIFAR100}} \\ \cline{4-5} 
 &  &  & \multicolumn{1}{c|}{\textbf{Accuracy (\%)}} & \textbf{Accuracy (\%)} \\ \hline
\multirow{9}{*}{6Conv+FC} & \multirow{7}{*}{128} & \textit{PSN} \cite{fang2023parallel} & \multicolumn{1}{c|}{88.45} & 62.21 \\  
 &  & \textit{masked PSN} \cite{fang2023parallel} & \multicolumn{1}{c|}{85.81} & 60.69 \\ 
 &  & \textit{GLIF} \cite{yao2022glif} & \multicolumn{1}{c|}{83.66} & 58.92 \\ 
 &  & \textit{KLIF} \cite{jiang2023klif} & \multicolumn{1}{c|}{83.26} & 57.37 \\ 
 &  & \textit{PLIF} \cite{fang2021incorporating} & \multicolumn{1}{c|}{83.49} & 57.55 \\ 
 &  & \textit{LIF} & \multicolumn{1}{c|}{81.50} & 55.45 \\ 
 &  & \textit{FLAMES} & \multicolumn{1}{c|}{90.25} & 65.33 \\ \cline{2-5} 
 & 64 & \textit{FLAMES} & \multicolumn{1}{c|}{\textbf{88.62}} & \textbf{63.57} \\ 
 & 32 & \textit{FLAMES} & \multicolumn{1}{c|}{\textbf{83.15}} & \textbf{56.32} \\ \hline
\end{tabular}%
}
\end{table}

% \section{New Results }

% \begin{table}[h]
% \centering
% \caption{\hlt{Results comparing the accuracy of our model against some spiking and non-spiking architectures on test sets of LRA benchmark tasks.}}
% \label{tab:lra}
% \resizebox{\columnwidth}{!}{%
% \hlt{
% \begin{tabular}{|l|c|c|c|c|c|c|}
% \hline
% \textbf{Model} & \textbf{SNN} & \textbf{ListOps} & \textbf{Text} & \textbf{Retrieval} & \textbf{Image} & \textbf{Pathfinder} \\ \hline
% S4 (Original) \cite{gu2022efficiently} & No & 58.35 & 76.02 & 87.09 & 87.26 & 86.05 \\ 
% S4 (Improved) \cite{gu2022efficiently} & No & 59.60 & 86.82 & 90.90 & 88.65 & 94.20 \\ 
% Transformer \cite{vaswani2017attention} & No & 36.37 & 64.27 & 57.46 & 42.44 & 71.40 \\ 
% Sparse Transformer \cite{tay2020long} & No & 17.07 & 63.58 & 59.59 & 44.24 & 71.71 \\ 
% Linformer \cite{wang2020linformer} & No & 35.70 & 53.94 & 52.27 & 38.56 & 76.34 \\ 
% Linear Transformer \cite{tay2020long} & No & 16.13 & 65.90 & 53.09 & 42.34 & 75.30 \\ 
% FLASH-quad \cite{hua2022flash} & No & 42.20 & 64.10 & 83.00 & 48.30 & 83.62 \\ 
% Spiking LMUFormer \cite{liu2024lmuformer} & Yes & 37.30 & 65.80 & 79.76 & 55.65 & 72.68 \\ 
% TransNormer T2 \cite{qin2022transnormer} & No & 41.60 & 72.20 & 83.82 & 49.60 & 76.60 \\ 
% BinaryS4D \cite{stan2023binary} & Partial & 54.80 & 82.50 & 85.30 & 82.00 & 82.60 \\ 
% \textbf{FLAMES (Our Model)} & Yes & \textbf{59.08} & \textbf{79.41} & \textbf{89.62} & \textbf{79.88} & \textbf{86.47} \\ \hline
% \end{tabular}%
% }
% }
% \end{table}

\subsection{DVS Gesture Recognition}

To further investigate the combined effectiveness of dendritic mechanisms and \textit{FLAMES} convolutions in event-based processing, we evaluate our model on the \textit{DVS Gesture} dataset. This dataset consists of event streams recorded from a Dynamic Vision Sensor (DVS) at a resolution of $128 \times 128$, providing a challenging benchmark for evaluating temporal dynamics in gesture recognition tasks involving varying speeds and motions.

Our goal is to assess how the integration of dendritic mechanisms with \textit{FLAMES} convolution layers enhances the model's ability to capture multi-scale temporal dependencies. Specifically, we examine how dendrites can serve as a temporal attention mechanism that helps \textit{FLAMES} effectively focus on the most relevant events, while \textit{FLAMES} convolutions manage the overall temporal and spatial evolution of features.

The experiment involves training variants of our model—one incorporating both dendritic mechanisms and \textit{FLAMES} convolutions, and the other using only \textit{FLAMES}—to determine the contribution of dendritic attention. Table \ref{tab:combined_comparison_dvs128_gesture} summarizes the test accuracy of our models compared to other state-of-the-art approaches. The results are measured in terms of classification accuracy, along with the number of parameters, to highlight model efficiency.

As shown in Table \ref{tab:combined_comparison_dvs128_gesture}, the \textit{FLAMES} model with 128 channels, incorporating dendritic attention, achieves 96.5\% accuracy while maintaining a significantly lower parameter count compared to many other state-of-the-art models. This shows that our approach effectively utilizes sparse event-driven inputs to achieve high accuracy with reduced computational complexity. The use of dendritic mechanisms allows the model to dynamically adjust its focus on different temporal scales, thus improving gesture recognition even in scenarios with rapid motion changes.

The variant without dendritic attention, while still competitive, lags behind in adapting to the multi-scale nature of the event data, especially for gestures with complex temporal characteristics. This indicates that the dendritic mechanism plays a crucial role in adaptively filtering relevant temporal features, which is essential for handling the asynchronous, irregular inputs typical of event cameras.

In addition, visualizations of the learned dendritic activity reveal how the model attends to different time segments, effectively filtering the incoming spike streams to prioritize the most relevant events. This adaptive filtering complements the \textit{FLAMES} convolutional operations, leading to more robust and efficient temporal feature extraction.

Overall, the results validate the utility of combining dendritic mechanisms with \textit{FLAMES} convolutions for event-driven tasks, making the model well-suited for gesture recognition from DVS inputs. The joint use of these components allows for efficient temporal modeling, maintaining a favorable trade-off between accuracy and parameter efficiency.

\subsection{Scaling to HD Event Streams}

\begin{table}{r}%{0.95\textwidth} % Wider table for better fit
\centering
\caption{Latency Comparison on Celex-HAR (in milliseconds)}
\begin{tabular}{@{}lc||lc||lc@{}}
\toprule
\textbf{Algorithm} & \textbf{Latency (ms)} & \textbf{Algorithm} & \textbf{Latency (ms)} & \textbf{Algorithm} & \textbf{Latency (ms)} \\ \midrule
\textbf{FLAMES-Tiny}     & \textbf{0.162}   & TAM                  & 76.012   & VideoMamba-S       & 19.707   \\
\textbf{FLAMES-Small}    & \textbf{0.582}   & GSF                  & 75.558   & VideoMamba-M       & 58.164   \\
\textbf{FLAMES-Normal}   & \textbf{1.867}   & V-SwinTrans          & 39.837   & EVMamba            & 170.34   \\
ResNet-50               & 41.575           & TimeSformer          & 255.425  & EVMamba w/o Voxel Scan & 82.423   \\
C3D                     & 0.473            & SlowFast             & 1.118    & VRWKV-S            & 21.091   \\
R2Plus1D                & 94.264           & EFV++                & 166.23   & VRWKV-B            & 86.346   \\
TSM                     & 1.4266           & ESTF                 & 80.61    & Vision Mamba-S     & 23.88    \\
ACTION-Net              & 81.035           & SVFormer             & 897.455  & VMamba-S           & 53.302   \\
TAM                     & 76.012           &                      &          & VMamba-S(V2)       & 39.848   \\
GSF                     & 75.558           &                      &          & VMamba-B           & 82.421   \\
V-SwinTrans             & 39.837           &                      &          & VMamba-B(V2)       & 70.514   \\
\bottomrule
\end{tabular}
\label{tab:latency_celex}
\end{table}

The scalability of the proposed \textit{FLAMES} model is evaluated on the \textit{Celex HAR} dataset, a human activity recognition dataset recorded at a high resolution of $1280 \times 800$. This dataset serves as a challenging benchmark for assessing the model's ability to maintain high accuracy and computational efficiency when processing large-scale spatial and temporal data.

In this experiment, \textit{FLAMES} is used for action recognition on HD event streams, and its performance is compared to that of baseline Spiking Neural Networks (SNNs) and State-Space Models (SSMs). As shown in Figure \ref{fig:flops}, the results demonstrate that \textit{FLAMES} maintains high accuracy even at increased resolutions, whereas the baseline models experience significant performance degradation due to heightened computational demands. The integration of the \textit{FLAMES convolution layer} proves effective in managing the complex spatial and temporal components of HD event data, providing robust real-time processing capabilities with minimal computational overhead.

Figure \ref{fig:flops} illustrates the trade-off between accuracy and computational cost, measured in terms of FLOPs, for our \textit{FLAMES} models compared to state-of-the-art methods on the \textit{Celex-HAR} dataset. The \textit{FLAMES} variants—\textit{FLAMES Tiny}, \textit{FLAMES Small}, and \textit{FLAMES Normal}—demonstrate superior efficiency by achieving competitive or better accuracy while utilizing significantly fewer computational resources.

Key observations from Figure \ref{fig:flops} are as follows: 

\begin{itemize} \item \textbf{Efficiency at Different Scales}: \textit{FLAMES Tiny} achieves approximately 63.8\% accuracy with a fraction of the computational cost compared to larger models such as \textit{SlowFast} and \textit{C3D}. As the model scales to \textit{FLAMES Small} and \textit{FLAMES Normal}, accuracy improves to 69.3\% and 72.1\%, respectively, while maintaining a favorable computational cost profile. \item \textbf{Performance with Reduced Complexity}: \textit{FLAMES Normal} matches or exceeds the accuracy of models like \textit{TSM} and \textit{VisionMamba-S} but at a substantially lower computational cost. This efficiency is attributed to the integration of event-driven processing and effective state-space dynamics. \end{itemize}

The improved efficiency of \textit{FLAMES} can be credited to the event-based processing capabilities of the \textbf{FLAMES architecture} and the \textbf{FLAMES convolution layer}, which optimally manage state-space evolution without relying on dense operations. These features allow the model to capture complex temporal dependencies while minimizing computational requirements, making \textit{FLAMES} particularly effective for high-resolution event-based datasets like \textit{Celex-HAR}.

\hlt{

\textbf{HAR-DVS Results: }The HAR-DVS dataset results underscore the advantages of our FLAMES models, achieving accuracies of \textbf{70.38\%}, \textbf{81.73\%}, and \textbf{88.29\%} for FLAMES-Tiny, FLAMES-Small, and FLAMES-Normal, respectively, while maintaining substantially lower computational costs compared to other state-of-the-art models. Unlike traditional deep neural networks such as C3D and R2Plus1D, which struggle to model the complex temporal relationships inherent in event streams, FLAMES leverages a novel \textit{event-by-event processing approach}, preserving fine-grained temporal dynamics essential for accurate action recognition.

Moreover, FLAMES employs a unique \textit{dendritic attention mechanism} that enhances its ability to capture long-range spatio-temporal dependencies efficiently. The prolonged and complex actions in HAR-DVS demand robust temporal attention mechanisms, as highlighted in prior studies. FLAMES's dendritic-inspired design meets these requirements while offering a computationally efficient solution, making it particularly suitable for real-time, low-latency applications in dynamic event-driven environments.

 It is important to note that HAR-DVS provides frame-based data, as raw event data was unavailable for download. Since FLAMES is designed for event-by-event processing, we treated all events arriving at the same timestamp as a single batch for processing, adhering to the event-driven principles of the model.

}

\newpage

\newpage
\section{Supplementary Section C: Methods and Architectural Details}
\label{sec:architecture}
% Add content here

\begin{algorithm}[h]
\caption{FLAMES Model Training}
\label{alg:FLAMES}
\resizebox{\columnwidth}{!}{ % Resize to fit within column width
\begin{minipage}{\columnwidth} % Allows resizing within a minipage
\begin{algorithmic}[1]
\Require Training dataset $\mathcal{D} = \{(\mathbf{X}_i, \mathbf{y}_i)\}_{i=1}^N$, learning rate $\eta$, total epochs $E$, threshold potential $V_{\text{th}}$, decay factors $\alpha_d$, $\beta$
\State Initialize weights $\mathbf{W}$, dendritic timing factors $\tau_d$, FLAMES matrices $\mathbf{A}, \mathbf{B}, \mathbf{C}$, \hlt{low-rank matrices $\mathbf{P}, \mathbf{Q}$, and kernel $\mathbf{K}(\omega)$}
\State Initialize coupling strengths $\mathbf{g}_d$ \hlt{for each dendrite $d$}

\For{epoch $= 1$ to $E$}
    \For{each $(\mathbf{X}, \mathbf{y}) \in \mathcal{D}$}
        \Statex \textbf{Input Representation:} Prepare input events for processing
        \State Parse input event sequence $\mathbf{X} = \{(x_i, y_i, t_i, p_i)\}$, where $(x_i, y_i)$ are spatial coords, $t_i$ is time, $p_i$ is polarity.
        
        \Statex \textbf{Dendrite Attention Layer:} Update dendritic currents and aggregate at soma
        \For{each $t_i$ in spike event sequence}
            \For{each dendrite $d$}
                \State Update dendritic current: $\mathbf{i}_d(t_i+1) = \alpha_d \cdot \mathbf{i}_d(t_i) + \sum_{j \in \mathcal{N}_d} \mathbf{w}_j \cdot p_j$
            \EndFor
            \State Aggregate currents at soma: $V(t_i+1) = \beta \cdot V(t_i) + \sum_{d} \mathbf{g}_d \cdot \mathbf{i}_d(t_i)$
            \If{$V(t_i+1) > V_{\text{th}}$}
                \State Generate spike and reset potential: $V(t_i+1) \gets 0$
            \EndIf
        \EndFor
        
        \Statex \textbf{Spatial Pooling Layer:} Reduce spatial dimensionality while preserving temporal resolution
        \State Apply max pooling: $I_{\text{pooled}}(x', y', t) = \max_{(x, y) \in P(x', y')} I(x, y, t)$
        
        \Statex \textbf{FLAMES Conv. Layer:} Apply SA-HiPPO, NPLR, \& FFT for event dynamics
        \State Initialize state vector $\mathbf{x}(0)$
        \For{each spike time $t_k$ in $I_{\text{pooled}}$}
            \State Compute $\Delta t_k = t_{k+1} - t_k$, decay $\mathbf{F}_{ij}(\Delta t_k) = e^{-\alpha_{ij} \cdot \Delta t_k}$
            \State Compute spike-aware HiPPO: $\mathbf{A_S} = \mathbf{A} \circ \mathbf{F}(\Delta t_k)$
            \State Decompose: $\mathbf{A_S} = \mathbf{V} \mathbf{\Lambda} \mathbf{V^*} - \mathbf{P} \mathbf{Q^*}$
            \State $\displaystyle e^{\mathbf{A_S} \Delta t_k} \approx \mathbf{I} + \mathbf{A_S} \Delta t_k + \frac{\mathbf{A_S}^2 (\Delta t_k)^2}{2}$
            \State Update: $\mathbf{x}(t_{k+1}) = e^{\mathbf{A_S} \Delta t_k} \cdot \mathbf{x}(t_k) + \mathbf{A_S}^{-1} (\mathbf{e}^{\mathbf{A_S} \Delta t_k} - \mathbf{I}) \cdot \mathbf{B} \cdot \mathbf{S}(t_k)$
            \State FFT-based convolution:
            $\mathbf{x}(t_{k+1}) = \text{IFFT}(\text{FFT}(\mathbf{K}(\omega)) \odot \text{FFT}(\mathbf{x}(t_{k+1})))$
        \EndFor
        \State Compute continuous output: $\mathbf{y}(t) = \mathbf{C} \cdot \mathbf{x}(t)$
        \State \textbf{Thresholding:} Convert $\mathbf{y}(t)$ to spikes by applying $y_{\text{spike}}(t) = \mathbb{I}(\mathbf{y}(t) > V_{\text{th}})$
        
        \Statex \textbf{Normalization:} Reduce variability in activations
        \State Apply layer normalization: $\hat{\mathbf{x}}_l = \frac{\mathbf{x}_l - \mu_l}{\sqrt{\sigma_l^2 + \epsilon}} \cdot \gamma + \beta$
    
        \Statex \textbf{Readout Layer:} Compute final output and update model parameters
        \State Compute pooled state: $\mathbf{x}_{\text{pooled}, k} = \frac{1}{p} \sum_{i=kp}^{(k+1)p - 1} \hat{\mathbf{x}}_i$
        \State Final output: $\mathbf{y}_{\text{pred}} = \mathbf{W} \cdot \mathbf{x}_{\text{pooled}} + \mathbf{b}$
        \State Compute loss $\mathcal{L}(\mathbf{y}_{\text{pred}}, \mathbf{y})$, update $\mathbf{W} \gets \mathbf{W} - \eta \cdot \frac{\partial \mathcal{L}}{\partial \mathbf{W}}$
    \EndFor
\EndFor
\end{algorithmic}
\end{minipage}
}
\end{algorithm}

\subsection*{Background and Preliminaries}

\textbf{State-Space Models: }A state-space model (SSM) is a mathematical framework for modeling systems that evolve over time. The dynamics of such systems are described by a set of first-order differential equations, often expressed in continuous time as:
\[
\dot{x}(t) = A x(t) + B u(t), \quad y(t) = C x(t) + D u(t)
\]
where:
\begin{itemize}
    \item \( x(t) \in \mathbb{R}^N \) is the hidden state vector, representing the internal state of the system at time \( t \),
    \item \( u(t) \in \mathbb{R}^M \) is the input signal, such as sensory data or external stimuli,
    \item \( y(t) \in \mathbb{R}^P \) is the output signal or observable state,
    \item \( A \in \mathbb{R}^{N \times N} \), \( B \in \mathbb{R}^{N \times M} \), \( C \in \mathbb{R}^{P \times N} \), and \( D \in \mathbb{R}^{P \times M} \) are learned system matrices.
\end{itemize}

State-space models are often used in signal processing and control systems to model systems with temporal dependencies. In many practical scenarios, however, the continuous-time formulation is discretized:
\[
x_{k+1} = A_d x_k + B_d u_k, \quad y_k = C_d x_k + D_d u_k
\]
where \( A_d \), \( B_d \), \( C_d \), and \( D_d \) are the corresponding discrete-time matrices, and \( k \) indexes the discrete time steps.

\textbf{Spiking Neural Networks (SNNs):} SNNs are a class of neural networks that more closely mimic biological neurons. In SNNs, information is transmitted as spikes, or binary events, at discrete times, as opposed to continuous activations in traditional neural networks. A typical neuron in an SNN, such as the \textit{Leaky Integrate-and-Fire (LIF)} neuron, is governed by the following dynamics:
\[
\tau_m \frac{dV_i(t)}{dt} = -V_i(t) + I_i(t)
\]
where \( V_i(t) \) is the membrane potential of neuron \( i \), \( \tau_m \) is the membrane time constant, and  \( I_i(t) \) is the input current, typically derived from presynaptic neurons or external stimuli.

A spike is emitted when the membrane potential exceeds a threshold \( \theta_i \). After a spike, the membrane potential is reset, and a refractory period prevents immediate re-firing. 

Despite their potential for efficient temporal data processing, SNNs are difficult to train due to the non-differentiability of spikes and the complex membrane potential dynamics.

\textbf{Highly Optimized Polynomial Projection (HiPPO): }The \textit{HiPPO} framework provides a method for approximating the continuous history of an input signal by projecting it onto a set of polynomial basis functions. The HiPPO matrix \( A \) is designed to optimally compress the history of the input into a state vector \( x(t) \), allowing the model to retain relevant temporal dependencies over long time scales. For example, the HiPPO-Legendre (HiPPO-LegS) matrix \( A \) is defined as:
\[
A_{nk} = \begin{cases} 
-\sqrt{(2n+1)(2k+1)} & \text{if} \ n > k \\
n + 1 & \text{if} \ n = k \\
0 & \text{if} \ n < k
\end{cases}
\]
This matrix governs the dynamics of how the internal state evolves to represent the history of the input in a compressed manner.

\hlt{

\subsection*{Mathematical Modeling and Spike Generation Mechanism}

% \subsubsection{Spike Generation and Propagation}
Spikes in the FLAMES model are generated through the dynamics of LIF neurons. The spike generation process is described in detail below:

\begin{itemize}
    \item \textbf{Dendritic Current Integration}: Each DH-LIF neuron integrates incoming spikes through its dendritic branches:
    \begin{equation}
        i_d(t+1) = \alpha_d i_d(t) + \sum_{j \in N_d} w_j p_j,
    \end{equation}
    where $\alpha_d = e^{-\frac{1}{\tau_d}}$ represents the decay rate, $w_j$ is the synaptic weight, and $p_j$ is the input spike value.

    \item \textbf{Soma Potential Update and Spike Generation}: The soma potential is updated based on the integrated dendritic currents:
    \begin{equation}
        V(t+1) = \beta V(t) + \sum_d g_d i_d(t),
    \end{equation}
    where $\beta = e^{-\frac{1}{\tau_s}}$ is the decay rate of the soma, and $g_d$ is the coupling strength of each dendrite. A spike is generated if $V(t)$ exceeds the threshold $V_{\text{th}}$.

    \item \textbf{Spike Propagation}: The generated spikes propagate through the network according to:
    \begin{equation}
        x(t_{k+1}) = e^{A \Delta t_k} x(t_k) + A^{-1} (e^{A \Delta t_k} - I) B S(t_k),
    \end{equation}
    preserving both spatial and temporal information.
\end{itemize}

}

\subsection*{Methods}

The proposed model is designed to handle sparse, asynchronous event-based inputs effectively while being scalable to high-definition (HD) event streams. It leverages \textit{Dendrite Heterogeneity Leaky Integrate-and-Fire (DH-LIF)} neurons in the first layer to capture \textit{multi-scale temporal dynamics}, crucial for preserving temporal details inherent in event streams while reducing spatial \hlt{ and computational }redundancy. The model then utilizes a series of \textit{spiking state-space convolution} layers, enabling efficient integration of both local and global temporal relationships. The final \textit{readout layer} employs event pooling and a linear transformation to produce a compact and meaningful representation for downstream tasks such as classification or regression. This architecture ensures robustness and scalability, making it suitable for high-resolution inputs.

\hlt{
\subsection*{Variables and Notations}
To ensure clarity, we provide definitions for all variables and notation used in the equations:

\begin{table*}[h]
\centering
\caption{Summary of Variables and Notations}
\resizebox{\textwidth}{!}{%
\begin{tabular}{@{}lc||lc||lc@{}}
\toprule
\textbf{Variable} & \textbf{Definition} & \textbf{Variable} & \textbf{Definition} & \textbf{Variable} & \textbf{Definition} \\ \midrule
\multicolumn{6}{c}{\textbf{Input Representation}} \\ \midrule
$x, y$ & Spatial coordinates of the spike event & $t$ & Timestamp of the spike & $p$ & Magnitude or polarity of the spike \\ \midrule
\multicolumn{6}{c}{\textbf{Dendrite Attention Layer}} \\ \midrule
$\tau_d$ & Dendritic timing factor & $i_d(t)$ & Dendritic current for branch $d$ at $t$ & $\alpha_d$ & Decay rate of dendrite $d$, $e^{-\frac{1}{\tau_d}}$ \\
$\mathcal{N}_d$ & Presynaptic inputs to dendrite $d$ & $w_j$ & Synaptic weight of presynaptic input $p_j$ & $V(t)$ & Soma membrane potential at $t$ \\
$\beta$ & Soma decay rate, $e^{-\frac{1}{\tau_s}}$ & $g_d$ & Coupling strength of dendrite $d$ & $V_{\text{th}}$ & Threshold potential for spike generation \\ \midrule
\multicolumn{6}{c}{\textbf{Spatial Pooling Layer}} \\ \midrule
$I(x,y,t)$ & Initial spike activity & $I_{\text{pooled}}(x',y',t)$ & Pooled spike activity & $P(x',y')$ & Pooling window center at $(x',y')$ \\ \midrule
\multicolumn{6}{c}{\textbf{FLAMES Convolution Layer}} \\ \midrule
$x(t)$ & Internal state vector at $t$ & $S(t)$ & Input spike train & $A_S$ & SA-HiPPO matrix for inter-spike intervals \\
$B, C$ & Input/output coupling matrices & $\Delta t$ & Inter-spike interval & $F(\Delta t)$ & SA-HiPPO decay matrix, $e^{-\alpha_{ij} \Delta t}$ \\
$V, \Lambda$ & Components of NPLR decomposition & $P, Q$ & Low-rank matrices, $r \ll N$ & $K(\omega)$ & FFT convolution kernel, $\frac{1}{\omega - \Lambda}$ \\
$\text{FFT}(\cdot)$ & Fast Fourier Transform & $\text{IFFT}(\cdot)$ & Inverse FFT &  &  \\ \midrule
\multicolumn{6}{c}{\textbf{Normalization Layer}} \\ \midrule
$x_l$ & Input at layer $l$ & $\mu_l, \sigma_l^2$ & Mean, variance at layer $l$ & $\gamma, \beta$ & Learnable scale and shift parameters \\ \midrule
\multicolumn{6}{c}{\textbf{Readout Layer}} \\ \midrule
$x_{\text{pooled}, k}$ & Pooled state vector & $W, b$ & Learnable weight matrix and bias & $y$ & Final output, $y = W x_{\text{pooled}} + b$ \\ 
\bottomrule
\end{tabular}
}
\label{tab:variables_notations}
\end{table*}

\subsection*{Overview of the FLAMES Model}
The proposed Spiking Network for Learning Long-Range Relations (FLAMES) addresses the limitations of conventional spiking neural networks (SNNs) in capturing long-range temporal dependencies while maintaining event-driven efficiency. The FLAMES model is composed of the following key components:

% \subsubsection*{Symbol Definitions}

\begin{algorithm}[H]
\caption{FLAMES Model Processing}
\label{alg:FLAMESProcessing}
\begin{algorithmic}[1]
\Require Input spike event sequence $X = \{(x_i, y_i, t_i, p_i)\}$
\State \textbf{Initialize} model parameters
\State \textbf{Process} input through \textbf{Dendrite Attention Layer} (Algorithm \ref{alg:DendriteAttention})
\State \textbf{Apply} \textbf{Spatial Pooling Layer} to reduce spatial dimensions (Algorithm \ref{alg:SpatialPooling})
\State \textbf{Pass} output to \textbf{FLAMES Convolution Layer} to capture temporal dynamics (Algorithm \ref{alg:FLAMESConvolution})
\State \textbf{Update} state using \textbf{Spike-Aware HiPPO} mechanism (Algorithm \ref{alg:FLAMESConvolution})
\State \textbf{Aggregate} information in the \textbf{Readout Layer} for final output (Algorithm \ref{alg:ReadoutLayer})
\State \textbf{Output}: Model prediction $y$
\end{algorithmic}
\end{algorithm}

}

\subsection{\textbf{Input Representation}}
The input to the model is represented as a sequence of spike events, each defined by the tuple $(x, y, t, p)$, where $(x, y)$ are the spatial coordinates, $t$ is the timestamp, and $p$ represents the magnitude or polarity of the spike. These events are streamed asynchronously, reflecting the sparse nature of the data. \hlt{The model is also designed to handle higher resolutions}, allowing scalability to HD event streams. This input representation emphasizes the need for efficient aggregation of both spatial and temporal information while minimizing computational load.

\subsection{\textbf{Dendrite Attention Layer}}
The model begins by passing the input through the \textit{Dendrite Attention Layer}, constructed using DH-LIF neurons as shown in Fig. \ref{fig:overall}. \hlt{Each DH-LIF neuron features multiple dendritic branches, each with a unique timing factor $\tau_d$, enabling the capture of temporal dynamics across a range of timescales}, which is essential for accommodating the diverse timescales present in asynchronous spike inputs. The dynamics of the dendritic current $i_d(t)$ are governed by $\displaystyle i_d(t+1) = \alpha_d i_d(t) + \sum_{j \in \mathcal{N}_d} w_j p_j$, where $\alpha_d = e^{-\frac{1}{\tau_d}}$ is the decay rate for branch $d$, and $w_j$ represents the synaptic weight associated with presynaptic input $p_j$. The set $\mathcal{N}_d$ represents the presynaptic inputs connected to dendrite $d$, ensuring that each dendrite captures temporal features independently, \hlt{functioning as independent temporal filters. Unlike a standard CUBA LIF neuron model, which integrates all inputs uniformly at the soma with a single timescale, the dendritic attention layer introduces multiple dendritic branches, each independently filtering inputs at different temporal scales. This design enables the neuron to selectively process asynchronous inputs and retain information across diverse temporal windows, providing greater flexibility and adaptability.}

The dendritic currents from each branch are aggregated at the soma, resulting in the membrane potential $\displaystyle V(t+1) = \beta V(t) + \sum_{d} g_d i_d(t)$, where $\beta = e^{-\frac{1}{\tau_s}}$ represents the soma’s decay rate, and $g_d$ represents the coupling strength of dendrite $d$ to the soma. A spike is generated whenever the membrane potential exceeds a threshold $V_{\text{th}}$, allowing the neuron to selectively fire only when sufficiently excited.

\begin{algorithm}[h]
\caption{Dendrite Attention Layer}
\label{alg:DendriteAttention}
\begin{algorithmic}[1]
\Require Input spike events $X = \{(x_i, y_i, t_i, p_i)\}$, dendritic timing factors $\{\tau_d\}$, synaptic weights $\{w_j\}$, coupling strengths $\{g_d\}$, threshold $V_{\text{th}}$
\State \textbf{Initialize} dendritic currents $i_d(0)$ and membrane potential $V(0)$
\For{each time step $t$}
    \For{each dendrite $d$}
        \State Compute decay rate: $\alpha_d \gets e^{-\frac{1}{\tau_d}}$
        \State Update dendritic current: $i_d(t+1) \gets \alpha_d i_d(t) + \sum_{j \in \mathcal{N}_d} w_j p_j$
    \EndFor
    \State Compute soma decay rate: $\beta \gets e^{-\frac{1}{\tau_s}}$
    \State Update membrane potential: $V(t+1) \gets \beta V(t) + \sum_{d} g_d i_d(t)$
    \If{$V(t+1) > V_{\text{th}}$}
        \State Generate spike at time $t+1$
        \State Reset membrane potential: $V(t+1) \gets 0$
    \EndIf
\EndFor
\State \textbf{Output}: Spatio-temporal features $I(x, y, t)$
\end{algorithmic}
\end{algorithm}

\subsection{\textbf{Spatial Pooling Layer}}
Following the dendritic attention layer, a \textit{Spatial Pooling Layer} is introduced to reduce the spatial dimensionality of the resulting output. Given the initial spike activity $I(x, y, t)$ at location $(x, y)$, the pooling operation reduces spatial dimensions while preserving temporal resolution:

\[
I_{\text{pooled}}(x', y', t) = \max_{(x, y) \in P(x', y')} I(x, y, t)
\]

where $P(x', y')$ is a pooling window centered at $(x', y')$. Pooling reduces spatial complexity, simplifying subsequent processing in the network while retaining key features. \hlt{This is especially useful for HD event streams with extensive spatial information.}

\begin{algorithm}[h]
\caption{Spatial Pooling Layer}
\label{alg:SpatialPooling}
\begin{algorithmic}[1]
\Require Input spike activity $I(x, y, t)$ from Dendrite Attention Layer, pooling window $P(x', y')$
\For{each spatial location $(x', y')$}
    \For{each time step $t$}
        \State Pool activity: $I_{\text{pooled}}(x', y', t) \gets \displaystyle \max_{(x, y) \in P(x', y')} I(x, y, t)$
    \EndFor
\EndFor
\State \textbf{Output}: Pooled spike activity $I_{\text{pooled}}(x', y', t)$
\end{algorithmic}
\end{algorithm}

\subsection{\textbf{FLAMES Convolution} }
\label{subsec:FLAMES}

\hlt{
The \textbf{Spiking Process with Long-term Recurrent dynamics (FLAMES) Convolution Layer} is a critical component of the FLAMES model, specifically designed for processing event-based spiking inputs. It captures long-range dependencies and asynchronous dynamics by integrating mechanisms such as the \textbf{Spike-Aware HiPPO (SA-HiPPO)} framework, \textbf{Normal Plus Low-Rank (NPLR) Decomposition}, and \textbf{Fast Fourier Transform (FFT) Convolution}. These innovations collectively enable efficient and robust temporal feature extraction.

\textbf{Overview and Intuition}

Traditional convolutional layers are adept at extracting spatial features but often fail to capture complex temporal dependencies, especially in asynchronous, sparse spiking data. The FLAMES Convolution Layer overcomes this limitation by incorporating state-space models that inherently manage temporal dynamics. Leveraging the SA-HiPPO mechanism, the layer dynamically adapts memory retention based on spike timings, emphasizing recent events while allowing older information to decay. The use of NPLR Decomposition and FFT-based convolution further enhances computational efficiency, enabling scalability to high-dimensional, long-range temporal data.

}
\textbf{Spiking State-Space Model:} The temporal dynamics of the FLAMES Convolution Layer are governed by the \textbf{Spiking State-Space Model}:
\begin{equation}
\dot{x}(t) = A_S x(t) + B S(t), \quad y(t) = C x(t),
\end{equation}
where:
\begin{itemize}
    \item \( x(t) \in \mathbb{R}^N \) represents the internal state vector,
    \item \( S(t) \in \mathbb{R}^M \) is the input spike train, with each component \( S_i(t) = \sum_k \delta(t - t_i^k) \), where \( \delta(t) \) is the Dirac delta function,
    \item \( A_S \in \mathbb{R}^{N \times N} \) is the \textbf{Spike-Aware HiPPO} matrix,
    \item \( B \in \mathbb{R}^{N \times M} \) and \( C \in \mathbb{R}^{P \times N} \) are the input and output coupling matrices.
\end{itemize}

This framework ensures that temporal dependencies inherent in spiking data are captured effectively.

\begin{figure}
  \begin{center}
    \includegraphics[width=0.6\textwidth]{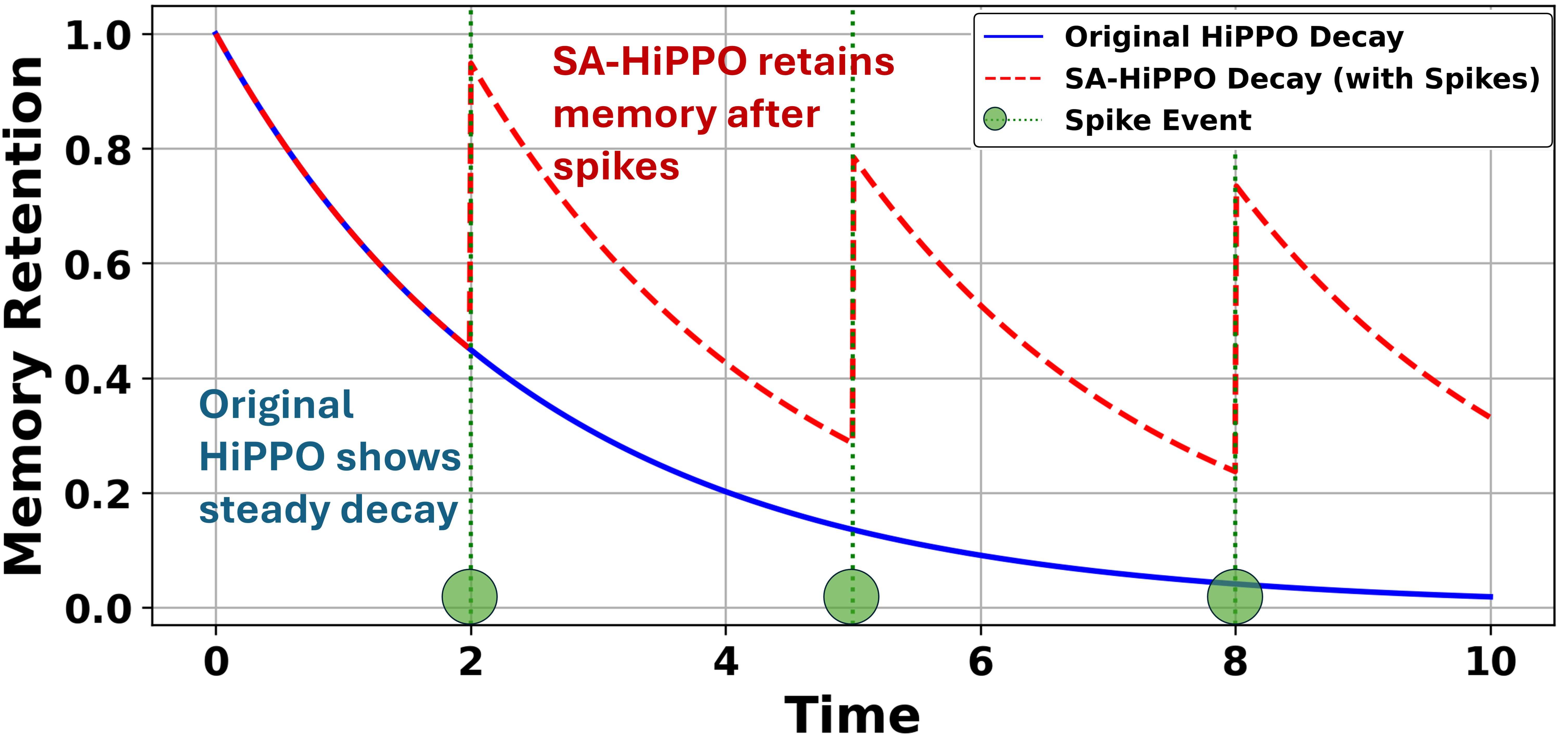}
  \end{center}
  \caption{The SA-HiPPO decay is needed to adapt the memory retention dynamically to the irregular timing of spike events, allowing the system to prioritize recent spikes while efficiently managing the decay of older information, which enhances stability and responsiveness for event-driven inputs.}
  \label{fig:sahippo}
\end{figure}

\textbf{Spike-Aware HiPPO Mechanism}: The \textit{Spike-Aware HiPPO (SA-HiPPO)} (Fig. \ref{fig:sahippo}) mechanism is a core component of the FLAMES model, designed to efficiently capture long-term temporal dependencies in the presence of sparse, event-based spiking inputs. The HiPPO (Highly Optimized Polynomial Projection) framework, originally developed to approximate continuous input signals, projects them onto polynomial bases, enabling efficient temporal compression of input history. However, when dealing with spike-driven dynamics, where inputs are discrete and irregular, the conventional HiPPO formulation must be adapted to properly address these challenges.
The SA-HiPPO adapts the HiPPO framework to efficiently handle discrete, spike-driven inputs by introducing a decay matrix \( F(\Delta t) \). This matrix adjusts memory retention based on the time elapsed between spikes (\( \Delta t \)), ensuring more recent spikes have a greater influence while older information gradually decays.
The Hadamard product with the original HiPPO matrix enables adaptive modulation of memory, making it more stable and suitable for asynchronous events.  In a spike-driven scenario, the input signal is represented as a vector of spike trains \( S(t) \in \mathbb{R}^M \), with each element \( S_i(t) \) defined by $\displaystyle S_i(t) = \sum_k \delta(t - t^k_i)$, where \( \delta(t) \) is the Dirac delta function, and \( t^k_i \) denotes the time of the \( k \)-th spike for input \( i \). Given the irregular and sparse nature of these spike-driven inputs, we introduce a \textit{Spike-Aware HiPPO (SA-HiPPO)} matrix \( A_S \) that extends the dynamics of the standard HiPPO to efficiently process spikes.
The SA-HiPPO matrix \( A_S \) modifies the original HiPPO dynamics to adapt to the nature of spiking events by incorporating a decay function that accounts for the time elapsed between successive spikes. Specifically, the state evolution in the presence of spikes is modeled by $\displaystyle \dot{x}(t) = A_S x(t) + B S(t)$. The matrix \( A_S \) is defined as $\displaystyle
A_S = A \circ F(\Delta t)$, where \( A \in \mathbb{R}^{N \times N} \) is the original HiPPO matrix, and \( F(\Delta t) \in \mathbb{R}^{N \times N} \) is a decay matrix that weights the original HiPPO dynamics based on the inter-spike interval \( \Delta t \). The operator \( \circ \) denotes the element-wise (Hadamard) product. The decay matrix \( F(\Delta t) \) is formulated as $\displaystyle F_{ij}(\Delta t) = e^{-\alpha_{ij} \Delta t}$, where \( \Delta t = t_j - t_i \) represents the time difference between spike \( i \) and spike \( j \), and \( \alpha_{ij} \) is a decay parameter that controls how the influence of past spikes diminishes over time. The exponential decay function ensures that the impact of previous spikes decreases exponentially, allowing more recent spikes to have a stronger influence on the current state. This weighting mechanism makes the HiPPO dynamics more adaptable to spiking inputs, capturing both the recency and relevance of spikes for efficient temporal representation.

The state vector \( x(t) \) thus evolves in two distinct modes: continuous evolution between spikes and instantaneous updates at spike times. Between spikes, the state evolves according to the homogeneous equation $\displaystyle \dot{x}(t) = A_S x(t)$. When a spike occurs at time \( t_k \), the state is updated as:

\[
x(t_{k+1}) = e^{A_S \Delta t_k} x(t_k) + A_S^{-1} \left( e^{A_S \Delta t_k} - I \right) B S(t_k)
\]

where \( \Delta t_k = t_{k+1} - t_k \) represents the time difference between successive spikes. To make the state update computationally feasible, the matrix exponential \( e^{A_S \Delta t_k} \) is approximated using a truncated Taylor series expansion:

\[
e^{A_S \Delta t_k} \approx I + A_S \Delta t_k + \frac{A_S^2 \Delta t_k^2}{2}
\]

This first-order or second-order approximation provides a good balance between computational efficiency and accuracy, especially in scenarios with small inter-spike intervals.

The SA-HiPPO mechanism effectively extends the temporal memory capabilities of the original HiPPO framework by introducing a spike-sensitive adaptation. It ensures that the state vector \( x(t) \) retains relevant temporal information while accommodating the asynchronous nature of spike inputs. The decay function embedded within \( F(\Delta t) \) provides a means to dynamically adjust the influence of past inputs, thereby making the model more responsive to recent events.

\textbf{Normal Plus Low-Rank (NPLR) Decomposition}: The \textbf{NPLR Decomposition} reduces computational complexity by expressing \( A_S \) as:
\begin{equation}
A_S = V \Lambda V^* - P Q^*,
\end{equation}
where:
\begin{itemize}
    \item \( V \in \mathbb{C}^{N \times N} \) is a unitary matrix,
    \item \( \Lambda \in \mathbb{C}^{N \times N} \) is a diagonal matrix of decay rates,
    \item \( P, Q \in \mathbb{C}^{N \times r} \) are low-rank matrices, with \( r \ll N \).
\end{itemize}

This decomposition reduces the complexity of matrix-vector multiplications from \( O(N^2) \) to \( O(Nr) \), facilitating scalability to large state spaces.

\textbf{Fast Fourier Transform (FFT) Convolution}: Long-range temporal dependencies are handled efficiently using FFT-based convolution. The convolution operation is performed as follows:
\begin{enumerate}
    \item Transform the state vector \( x(t) \) and convolution kernel \( K(\omega) \) into the frequency domain using FFT.
    \item Perform element-wise multiplication in the frequency domain.
    \item Apply the inverse FFT (IFFT) to obtain the updated state vector in the time domain.
\end{enumerate}

This approach significantly accelerates the processing of long temporal sequences by leveraging frequency-domain efficiencies. The FLAMES Convolution Layer integrates these components to achieve robust spatio-temporal feature extraction:
\begin{itemize}
    \item \textbf{Temporal Dynamics Modeling}: SA-HiPPO captures spike timing dependencies while balancing memory retention and decay.
    \item \textbf{Computational Efficiency}: NPLR Decomposition and FFT convolution ensure scalability and rapid processing.
    \item \textbf{Efficient State Management}: The state-space formulation ensures accurate updates for spiking inputs.
\end{itemize}

\subsubsection{FLAMES Convolution Layer}
\label{subsubsec:FLAMESconv}
\hlt{
Using all these concepts of SA-Hippo, NPLR Decomposition and FFT Convolution, we introduce \textit{FLAMES Convolution (FLAMESConv)} layers, which generalize the spike-aware state-space operations into a convolutional framework. These layers are designed to extend the capabilities of FLAMES by transforming the temporal memory operations into a convolutional form, thus allowing for more efficient feature extraction in both temporal and spatial domains.
The \textit{FLAMES Conv} layer incorporates spike-based input while retaining the convolutional structure, enabling the model to operate efficiently over high-dimensional data while capturing complex temporal dependencies. The continuous-time state-space dynamics are given by:

\[
\frac{d}{dt} x(t) = A x(t) + B u(t)
\]

where $x(t) \in \mathbb{R}^N$ represents the state vector, $u(t) \in \mathbb{R}^M$ is the input, $A \in \mathbb{R}^{N \times N}$ is the state transition matrix, and $B \in \mathbb{R}^{N \times M}$ is the input coupling matrix. The state evolves based on both the internal dynamics and the influence of incoming spikes. The Spike-Aware dynamics incorporate both decay and event-driven updates 
\begin{equation}\dot{x}(t) = A_{\text{spike}}(t) x(t) + B_{\text{spike}}(t) u(t),\end{equation} where $\displaystyle A_{\text{spike}}(t) = A_{\text{decay}} + A_{\text{timing}}(t)$. The matrix $A_{\text{decay}} = -\frac{1}{\tau_m} I$ models natural decay, while $A_{\text{timing}}(t)$ represents spike-driven effects and depends on the inter-spike intervals.
The model discretizes these dynamics for efficient implementation, using a fixed time step $\Delta t$: \begin{equation} x_{k+1} = x_k + \Delta t (A_{\text{spike}, k} x_k + B_{\text{spike}, k} u_k)\end{equation}

At each spike time $t_i$, the state undergoes an instantaneous update $\displaystyle
x(t_i^+) = x(t_i^-) + B_{\text{spike}}(t_i)$. To improve computational efficiency, the spiking state matrix $A_{\text{spike}}$ is decomposed using the \textit{Normal Plus Low-Rank (NPLR) decomposition}:$\displaystyle A_{\text{spike}} = V \Lambda V^* - P Q^*$

where $V \in \mathbb{C}^{N \times N}$ is a unitary matrix, $\Lambda \in \mathbb{C}^{N \times N}$ represents the decay, and $P, Q \in \mathbb{C}^{N \times r}$ are low-rank matrices. This reduces the cost of matrix-vector products from $O(N^2)$ to $O(Nr)$, where $r$ is the rank of the low-rank perturbation. The resulting state update rule becomes:

\[
x_{k+1} = x_k + \Delta t \left( (V \Lambda V^* - P Q^*) x_k + B_{\text{spike}} u_k \right)
\]

The convolution operation in these layers is realized by transforming recurrent state-space updates into a convolutional form, with the system's impulse response precomputed. Using the \textit{Fast Fourier Transform (FFT)}, the convolution kernel $K(\omega)$ can be efficiently calculated as $\displaystyle
K(\omega) = \frac{1}{\omega - \Lambda}$. This transformation allows the model to handle long-range temporal dependencies efficiently, even in high-resolution event-based streams.

\textbf{Computational Efficiency: }The layer achieves notable computational advantages:
\begin{itemize}
    \item \textbf{Reduced Complexity}: NPLR Decomposition transforms operations from \( O(N^2) \) to \( O(Nr) \).
    \item \textbf{Accelerated Convolutions}: FFT convolution rapidly processes long temporal sequences.
    \item \textbf{Parallelization}: FFT operations are well-suited for parallel hardware architectures, enhancing performance.
\end{itemize}

\textbf{Spike Generation in FLAMES Convolution Layers}: Spikes in the FLAMES model are generated through the interaction of dendritic and soma compartments in the DH-LIF neurons. These neurons are integral to the Dendrite Attention Layer, which precedes each FLAMES convolution layer, ensuring asynchronous and event-driven signal processing. 

The dendritic branches act as independent temporal filters, accumulating and processing inputs over time:
\[
i_d(t+1) = \alpha_d i_d(t) + \sum_{j \in \mathcal{N}_d} w_j p_j,
\]
where \( \alpha_d = e^{-\frac{1}{\tau_d}} \) is the decay rate determined by the dendritic branch’s time constant \( \tau_d \), \( w_j \) is the synaptic weight, and \( p_j \) is the presynaptic spike.

The soma aggregates these currents, with its membrane potential evolving as:
\[
V(t+1) = \beta V(t) + \sum_d g_d i_d(t),
\]
where \( \beta = e^{-\frac{1}{\tau_s}} \) represents the soma’s decay factor, and \( g_d \) is the coupling strength of each dendrite \( d \). 

A spike is produced when the soma's membrane potential \( V(t) \) exceeds the threshold \( V_{\text{th}} \). After firing, the potential resets, and these spikes serve as inputs to the next FLAMES convolution layer. This mechanism ensures the model maintains its asynchronous event-driven processing nature while enabling precise temporal modeling across layers.

The \textbf{FLAMES Convolution Layer} combines the strengths of SA-HiPPO, NPLR Decomposition, and FFT Convolution to process asynchronous spiking inputs effectively. This integration enables the model to extract meaningful spatio-temporal features while maintaining computational efficiency and scalability, making it ideal for high-resolution, real-world applications.
}

\begin{algorithm}[h]
\caption{FLAMES Convolution Layer}
\label{alg:FLAMESConvolution}
\begin{algorithmic}[1]
\Require Spike train input $S(t)$, HiPPO base matrix $\mathbf{A}$, input coupling matrix $\mathbf{B}$, output coupling matrix $\mathbf{C}$, decay function $\mathbf{F}(\Delta t)$, time step $\Delta t$, low-rank matrices $\mathbf{P}$, $\mathbf{Q}$, total time $T$, rank $r$, state space dimension $N$, FFT convolution kernel $\mathbf{K}(\omega)$, threshold potential $V_{\text{th}}$
\Ensure Output spike map $Y_{\text{spike}}(t)$

\Statex \textbf{Initialization}
\State Initialize state vector $\mathbf{x} \gets 0$ \hfill \textit{($N$-dimensional state vector)}
\State Initialize output $Y_{\text{spike}} \gets []$ \hfill \textit{(Empty list to store spike outputs)}

\Statex \textbf{Precomputations}
\State Compute spike-aware HiPPO matrix: $\mathbf{A}_{\text{spike}} \gets \mathbf{A} \circ \mathbf{F}(\Delta t)$ \hfill \textit{(Hadamard product with decay function)}
\State Perform eigendecomposition: $\mathbf{V}, \mathbf{\Lambda} \gets \text{eig}(\mathbf{A}_{\text{spike}})$
\State Decompose using NPLR: $\mathbf{A}_{\text{NPLR}} \gets \mathbf{V} \mathbf{\Lambda} \mathbf{V}^* - \mathbf{P} \mathbf{Q}^*$

\For{$t = 1$ \textbf{to} $T$}
    \Statex \textbf{Spike-Driven Dynamics}
    \If{$S(t)$ contains spikes}
        \State Compute time difference: $\Delta t_k = t_{k+1} - t_k$
        \State Approximate matrix exponential:
        \[
        e^{\mathbf{A}_{\text{spike}} \Delta t_k} \approx \mathbf{I} + \mathbf{A}_{\text{spike}} \Delta t_k + \frac{(\mathbf{A}_{\text{spike}})^2 (\Delta t_k)^2}{2}
        \]
        \State Update state vector:
        \[
        \mathbf{x}(t_{k+1}) \gets \mathbf{x}(t_k) + \Delta t_k \left( (\mathbf{V} \mathbf{\Lambda} \mathbf{V}^* - \mathbf{P} \mathbf{Q}^*) \mathbf{x}(t_k) + \mathbf{B} S(t_k) \right)
        \]
    \Else
        \State Update state for continuous dynamics: $\mathbf{x} \gets e^{\mathbf{A}_{\text{spike}} \Delta t} \mathbf{x}$
    \EndIf

    \Statex \textbf{FFT-Based Convolution for Temporal Dependencies}
    \State Transform state and kernel to frequency domain:
    \[
    \mathbf{X}_{\text{freq}} \gets \text{FFT}(\mathbf{x}), \quad \mathbf{K}_{\text{freq}} \gets \text{FFT}(\mathbf{K}(\omega))
    \]
    \State Perform element-wise multiplication in frequency domain:
    \[
    \mathbf{Y}_{\text{freq}} \gets \mathbf{X}_{\text{freq}} \cdot \mathbf{K}_{\text{freq}}
    \]
    \State Transform back to time domain:
    \[
    \mathbf{x}(t_{k+1}) \gets \text{IFFT}(\mathbf{Y}_{\text{freq}})
    \]

    \State Compute continuous output: $y_t \gets \mathbf{C} \cdot \mathbf{x}(t)$
    \State Threshold the output to generate spikes:
    \[
    y_{\text{spike}}(t) \gets \mathbb{I}(y_t > V_{\text{th}})
    \]
    \State Append $y_{\text{spike}}(t)$ to $Y_{\text{spike}}$
\EndFor

\Statex
\textbf{Output:} $Y_{\text{spike}}$, the final spike map
\end{algorithmic}
\end{algorithm}

\begin{algorithm}[h]
\caption{Readout Layer}
\label{alg:ReadoutLayer}
\begin{algorithmic}[1]
\Require State vectors $\{x(t)\}$, pooling factor $p$, weights $W$, bias $b$
\For{each pooled time step $k$}
    \State Compute pooled state:
    \[
    x_{\text{pooled}, k} \gets \frac{1}{p} \sum_{i=kp}^{(k+1)p-1} x(t_i)
    \]
\EndFor
\State Compute final output:
\[
y \gets W x_{\text{pooled}} + b
\]
\State \textbf{Output}: Model prediction $y$
\end{algorithmic}
\end{algorithm}

\subsection{\textbf{Normalization and Residual}}
To maintain stability and ensure efficient learning, \textit{Layer Normalization (LN)} is applied after each spiking SSM convolution layer: $\displaystyle \hat{x}_l = \frac{x_l - \mu_l}{\sqrt{\sigma_l^2 + \epsilon}} \cdot \gamma + \beta$, where $\mu_l$ and $\sigma_l^2$ are the mean and variance of activations at layer $l$, respectively, and $\gamma, \beta$ are learnable parameters. Normalization reduces variability in activations, providing stable training regardless of fluctuations in inputs.

Additionally, \textit{residual connections} help propagate information across layers by defining $\displaystyle x_{l+1} = f(x_l) + x_l$, where $f(x_l)$ represents the transformation applied by the spiking convolution at layer $l$. Residual connections prevent vanishing gradients, allow lower-level feature retention, and enhance learning efficiency.

% \subsection{Stack of Spiking SSM Convolution Layers}
% The stack of spiking SSM convolution layers incrementally extracts higher-order spatiotemporal features from the input representation. As spikes propagate through these layers, the model reduces redundancy in temporal features, effectively compressing the data while retaining essential information. The state evolution for each layer is given by $\displaystyle x_{l+1}(t) = A x_l(t) + B u_l(t)$. By stacking these layers, the network progressively captures complex dependencies, leading to richer feature extraction at each subsequent layer.

\begin{table}[h]
\centering
\caption{Input-Output Descriptions for Each Block in the FLAMES Model}
\label{tab:input_output}
\resizebox{\textwidth}{!}{
\begin{tabular}{@{}l l l@{}}
\toprule
\textbf{Block} & \textbf{Input} & \textbf{Output} \\ 
\midrule

\multirow{1}{*}{\textbf{Input Representation}}   
& Spike events \((x, y, t, p)\): \((x, y)\) (spatial), \(t\) (time), \(p\) (magnitude/polarity) & Preprocessed spike events for subsequent layers \\ 

\midrule

\multirow{4}{*}{\textbf{Dendrite Attention Layer}} 
& Spike event stream with spatial and temporal coordinates \((x, y, t, p)\) & Aggregated membrane potential \(\mathbf{v}(t)\), capturing spatio-temporal features at multiple timescales. \\ 
\cmidrule(lr){2-3}
& \textit{Dendritic Current Update:} \\
& Previous dendritic current \(\mathbf{i}_d(t)\), synaptic weights \(\mathbf{w}_j\), and decay factor \(\alpha_d\) & Updated dendritic current \(\mathbf{i}_d(t+1) = \alpha_d \mathbf{i}_d(t) + \sum_{j \in \mathcal{N}_d} \mathbf{w}_j p_j\) \\
\cmidrule(lr){2-3}
& \textit{Soma Aggregation:} & Aggregated membrane potential \(\mathbf{v}(t+1) = \beta \mathbf{v}(t) + \sum_{d} \mathbf{g}_d \mathbf{i}_d(t)\) \\
& Inputs from dendritic currents \(\mathbf{i}_d(t+1)\), soma decay factor \(\beta\), and coupling strengths \(\mathbf{g}_d\) & \\
\cmidrule(lr){2-3}
& \textit{Spike Generation:} & Spike output if \(\mathbf{v}(t+1) > V_{\text{th}}\), and reset potential (\(\mathbf{v}(t+1) \gets 0\)) \\ 
\midrule

\textbf{Spatial Pooling Layer} & Aggregated spikes \(\mathbf{I}(x, y, t)\) from the Dendrite Attention Layer & Pooled spatio-temporal representation \(\mathbf{I}_{\text{pooled}}(x', y', t)\), with reduced spatial dimensions \\ 

\midrule

\multirow{4}{*}{\textbf{FLAMES Convolution Layer}} 
& Pooled spike features \(\mathbf{I}_{\text{pooled}}(x', y', t)\) & Processed state \(\mathbf{y}(t)\), thresholded to generate spikes. \\ 
\cmidrule(lr){2-3}
& \textit{SA-HiPPO:} Spike features and inter-spike intervals (\(\Delta t\)) & Adjusted state-space matrix \(\mathbf{A_S}\), incorporating memory retention through a decay matrix \\
\cmidrule(lr){2-3}
& \textit{NPLR Decomposition:} & Decomposed matrix \(\mathbf{A_S} = \mathbf{V} \mathbf{\Lambda} \mathbf{V}^* - \mathbf{P} \mathbf{Q}^*\), reducing computational complexity \\
& Adjusted state-space matrix \(\mathbf{A_S}\) & \\
\cmidrule(lr){2-3}
& \textit{Matrix Exponential Approximation:} & Approximated exponential \(e^{\mathbf{A_S} \Delta t_k}\) for efficient state updates \\
& Decomposed state-space matrix \(\mathbf{A_S}\), time step \(\Delta t_k\) & \\
\cmidrule(lr){2-3}
& \textit{FFT Convolution:} State vector \(\mathbf{x}(t_k)\) and precomputed impulse response \(\mathbf{K}(\omega)\) & Updated state vector \(\mathbf{x}(t_{k+1})\) after efficient frequency-domain convolution \\

\midrule

\textbf{Layer Normalization} & Intermediate activations \(\mathbf{x}_l\) from the FLAMES Convolution Layer & Normalized activations \(\hat{\mathbf{x}}_l\), ensuring stable training by reducing variability in activations \\ 

\midrule

\textbf{Readout Layer} & Normalized features \(\hat{\mathbf{x}}_l\) & Final output \(\mathbf{y}\), generated via event pooling and a linear transformation \\ 

\bottomrule
\end{tabular}
}
\end{table}

\subsection{ \textbf{Readout Layer}}
The readout layer is inspired by the \textit{Event-SSM} architecture and employs an \textit{event-pooling mechanism} to subsample the temporal sequence length. The pooled output is computed as $\displaystyle x_{\text{pooled}, k} = \frac{1}{p} \sum_{i=kp}^{(k+1)p-1} x_i$, where $p$ is the pooling factor. This operation ensures only the most relevant temporal features are retained, reducing computational burden while preserving key information. The resulting pooled sequence is passed through a linear transformation as $\displaystyle y = W x_{\text{pooled}} + b$ where $W$ and $b$ are learnable parameters. The combination of event pooling and linear transformation provides an efficient means for deriving a final representation suitable for downstream tasks, maintaining scalability even with longer event sequences.

\end{document}